\documentclass{article}

\usepackage{microtype}
\usepackage{graphicx}
\usepackage{booktabs} 
\usepackage{caption}

\usepackage[accepted]{icml2025}

\usepackage{amsmath}
\usepackage{amssymb}
\usepackage{mathtools}
\usepackage{amsthm}
\usepackage{subcaption}

\usepackage[capitalize,noabbrev]{cleveref}

\theoremstyle{plain}
\newtheorem{theorem}{Theorem}[section]

\newtheorem{lemma}[theorem]{Lemma}

\theoremstyle{definition}
\newtheorem{definition}[theorem]{Definition}

\theoremstyle{remark}
\newtheorem{remark}[theorem]{Remark}

\usepackage[textsize=tiny]{todonotes}

\begin{document}

\twocolumn[
\icmltitle{Demystifying MPNNs: Message Passing as Merely Efficient Matrix Multiplication}




\begin{icmlauthorlist}
\icmlauthor{Qin Jiang}{1}
\icmlauthor{Chengjia Wang}{1}
\icmlauthor{Michael Lones}{1}
\icmlauthor{Wei Pang}{1}

\end{icmlauthorlist}

\icmlaffiliation{1}{Department of Computer Science, University of Heriot-Watt, Edinburgh, UK}

\icmlcorrespondingauthor{Wei Pang}{W.Pang@hw.ac.uk}

\icmlkeywords{Machine Learning, ICML}

\vskip 0.3in
]



\printAffiliationsAndNotice{}  

\begin{abstract}

While Graph Neural Networks (GNNs) have achieved remarkable success, their design largely relies on empirical intuition rather than theoretical understanding. In this paper, we present a comprehensive analysis of GNN behavior through three fundamental aspects: 
(1) we establish that \textbf{$k$-layer} Message Passing Neural Networks efficiently aggregate \textbf{$k$-hop} neighborhood information through iterative computation, 
(2) analyze how different loop structures influence neighborhood computation, and (3) examine behavior across structure-feature hybrid and structure-only tasks. 
For deeper GNNs, we demonstrate that gradient-related issues, rather than just over-smoothing, can significantly impact performance in sparse graphs. We also analyze how different normalization schemes affect model performance and how GNNs make predictions with uniform node features, providing a theoretical framework that bridges the gap between empirical success and theoretical understanding.

\end{abstract}
\section{Introduction}

Graph Neural Networks (GNNs) are considered to be powerful in learning on graph-structured data, particularly through their iterative neighbor aggregation mechanism. 

Despite their widespread adoption as feature extractors for graph data, fundamental questions about GNNs' representational capabilities remain open \cite{dehmamyUnderstandingRepresentationPower2019}. The design of new GNN architectures often relies on empirical intuition and heuristics rather than theoretical foundations \cite{xuHowPowerfulAre2019}.

While GNNs integrate both structural and feature information for predictions, our understanding of how these components interact and influence the final predictions remains limited. It is commonly assumed that a $k$-layer GNN effectively synthesizes both structural and feature information by aggregating data from progressively larger neighborhoods. However, our research reveals a more nuanced reality: when increasing from $k$ to $(k+1)$ layers, the layer-wise iterative aggregation process effectively substitutes information from $k$-hop neighbors with that of $(k+1)$-hop neighbors, rather than building a cumulative representation as previously thought. This is because graph loops lead to the coexistence of multi-hop neighbors in $k$-hop neighbors.

This paper demystifies Message Passing Neural Networks (MPNNs) by revealing their fundamental nature: the message passing process is, at its core, a memory-efficient implementation of matrix multiplication operations. Through this lens, we demonstrate three key insights:

(1) In Section \ref{sc:k_layer},  we establish that a $k$-layer MPNN transforms node representations by iteratively aggregating information from $k$-hop neighborhoods. 

More precisely, we prove an approximation equivalence: a $k$-layer MPNN operating with adjacency matrix $A$ is approximately equivalent to a single-layer MPNN operating with adjacency matrix $A^k$. 

This result not only provides a formal characterization of how message passing depth relates to neighborhood influence in GNNs, but also reveals a computational advantage: while direct computation of $A^k$ requires storing the full power matrix and can exceed memory constraints for large graphs, the iterative message passing in GNNs achieves equivalent neighborhood aggregation through memory-efficient layer-wise operations.

(2) In Section \ref{loops}, we analyze how different types of graph loops affect $k$-hop neighborhood computation, as loops create additional paths between nodes and thus increase the density of $k$-hop neighborhoods.

(3) Finally, in Section \ref{sc:dichotomy}, we examine MPNN behavior across structure-feature hybrid tasks and structure-only tasks, revealing their underlying similarity: structure-only tasks are essentially structure-feature hybrid tasks where node degrees serve as the node features.

We challenge the conventional wisdom about deeper GNNs' performance degradation: contrary to the common over-smoothing \cite{ruschSurveyOversmoothingGraph2023} explanation, we experimentally demonstrate that gradient-related issues can be the primary cause for sparse graphs. In addition, we explain how GNNs predict with uniform features and how different normalization schemes fundamentally influence their performance.

In summary, our work provides a theoretical foundation for understanding GNN behavior through three key aspects: the relationship between network depth and neighborhood aggregation, the impact of graph loop structures, and the role of gradients in deep architectures, normalization influence. These theoretical insights not only bridge the gap between empirical success and mathematical understanding but also provide practical guidance for GNN architecture design and deployment across various applications.

The code for the experiments conducted in this paper is available at https://anonymous.4open.science/status/demystify-B30E. 

\paragraph{Notation and definitions}
A graph $G=(A,X)$ is a set of $N$ nodes connected via a set of edges. The adjacency matrix of a graph $A$ encodes graph topology, where each element $A_{ij}$ represents an edge from node $i$ to node $j$. In this paper, edges are directed, the undirected graph is considered to be a special case of directed graph where all edges have their reversed edges in the graph.
Each node \(i\) is assigned a feature vector \(\mathbf{x}_i \in \mathbb{R}^d\), and all the feature vectors are stacked to a feature matrix \(\mathbf{X} \in \mathbb{R}^{n \times d}\), where \(n\) is the number of nodes in \(G\). The set of neighbors of node \(i\) is denoted by \(\mathcal{N}(i)\).

We use $AB$ or $A \cdot B$ to denote the matrix product of matrices $A$ and $B$. 
All multiplications and exponentiations are matrix products, unless explicitly stated.
Lower indices $A_{ij}$ denote $i$, $j$th elements of $A$, and $A_i$ means the $i$th row.
$A^p$ denotes the $p$th matrix power of $A$.

\section{$k$-layer GNNs}
\label{sc:k_layer}

\subsection{$k$-order features}

\begin{definition}
    The \textbf{$k$-hop neighbor} of a node $v$ in a graph $G = (V, E)$ is any node $u \in V$ such that there is a directed path of $k$ consecutive edges from node $u$ to node $v$.
\end{definition}

\begin{definition}
    A $k$th order node feature, defined as $A^kX$, represents the result of multiplying the adjacency matrix $A$ with itself p times and then multiplying with the node feature matrix $X$.
    Particularly, $0$th order node feature is the original node feature.
\end{definition}

\begin{lemma}
For a graph $G=(V,E)$ with adjacency matrix A and node feature matrix X, the features aggregated from p-hop neighbors of each node are equivalent to the $k$th order node feature $A^kX$.
\end{lemma}

\begin{remark}
$A^kXW$ is a linear transformation of $k$-hop neighbor features $A^kX$ using weight matrix $W$.
\end{remark}

\begin{lemma}
\label{lm:Ap>0}
In the \( k \)-th power of the adjacency matrix \( A^k \), a non-zero element \( A^k_{ij} > 0 \) indicates that there exists at least one directed path of length exactly \( k \) from node \( i \) to node \( j \). Furthermore, the value of \( A^k_{ij} \) represents the total number of such paths.
\end{lemma}
The proof is provided in Appendix \ref{pf:Ap}.

\begin{figure}
    \centering
    \includegraphics[width=0.8\linewidth]{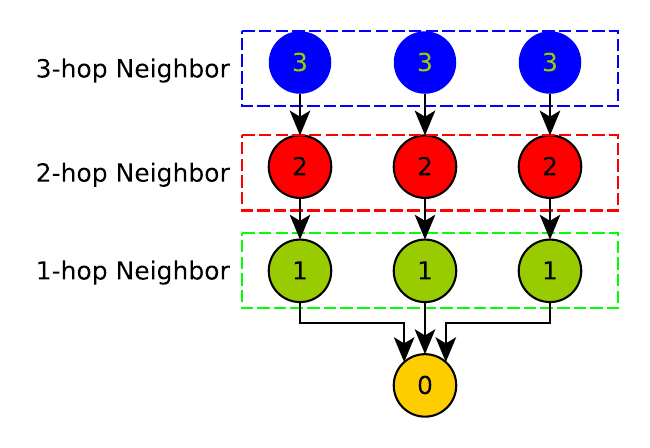}
    \caption{A $k$-layer GCN without adding selfloop will only gather information from $k$-hop neibors.}
    \label{fig:k_hop}
\end{figure}

\begin{remark}
The $k$th order node feature gathers information from nodes which are exactly $k$-hop away from the center node, as illustrated in Fig. \ref{fig:k_hop}.
\end{remark}

\subsection{Node representation of $k$-layer GNNs}

\begin{lemma}
\label{lm:gcn_no_loop}
For all natural numbers \(k\), the output of a \(k\)-layer GCN without self-loops can be expressed as:
\begin{equation}
  H^{(k)}  = \sigma\big((W \odot A)^k XW^{(k)} \big)
\end{equation}
\end{lemma}
The proof is provided in Appendix \ref{prf:gcn_no_loop}.

\begin{lemma}
\label{lm:gcn_loop}
For all natural numbers \(k\), the output of a \(k\)-layer GCN with self-loops can be expressed as:
\begin{equation}
      H^{(k)}  = \sigma\big((W \odot (A+I))^k XW^{(k)} \big)
\end{equation}
\end{lemma}
The proof is provided in Appendix \ref{proof:gcn-loop}.

$(A+I)^k$ can indeed be decomposed into a linear combination of powers of $A$, as described by the binomial theorem:
\[
(A + I)^k = \sum_{k=0}^k \binom{n}{k} A^k
\]
Where  \( \binom{n}{k} \) is the binomial coefficient. 

Therefore, the final representation \( H^{(k)} \) is a linear combination of the feature transformations derived from paths of lengths ranging from \( 0 \)-th to \( k \)-th order, capturing information aggregated over different scales in the graph.

\begin{lemma}
\label{lm:sage}
For all natural numbers \(k\), the output of a \(k\)-layer GraphSAGE can be expressed as:
\begin{equation}
\resizebox{\linewidth}{!}{
  $H^{(k)}  = \sigma\big((W \odot A)^k XW^{(k)}_n +...+(W \odot A) XW^{n}_1+ XW^{(k)}_0 \big)$}
\end{equation}
\end{lemma}
The proof is provided in Appendix \ref{prf:sage}.

The final representation of a $k$-layer GraphSAGE, similar to a GCN with self-loops, is derived from a combination of linear transformations applied to graph features aggregated from $0$-th to $k$-th order neighborhoods.

\subsection{Summary of $k$-layer GNNs}
\label{sc:sum_k_uat}

In summary, for a $k$-layer GNN, both GCN with self-loops and GraphSAGE integrate information from all neighborhood orders up to $k$. In contrast, a GCN without self-loops incorporates information solely from the $k$-th order neighborhood, as lower-order features are excluded in the absence of self-loops.

The approximation capabilities of graph neural networks (GNNs) reveal that a $k$-layer GNN with an adjacency matrix $A$ has the same approximation power as a 1-layer GCN with the adjacency matrix $A^k$. This observation demystifies the iterative aggregation power of message-passing neural networks (MPNNs).

In essence, multiple iterations of aggregation are equivalent to performing high-order matrix multiplications.

However, adding self-loops (as in GCNs with self-loops) or concatenating self-node features (as in GraphSAGE) incorporates features of all orders. While this can enhance the expressiveness of the model, it may also lead to over-smoothing, ultimately limiting the depth of GNNs and their ability to capture meaningful representations in deeper architectures.

\section{Loops}
\label{loops}

In Section \ref{sc:k_layer}, we discussed the influence of self-loops in GCNs. In this section, we will extend our discussion to consider all types of loops in graph neural networks and analyze their effects.

\begin{figure}
    \centering
    \includegraphics[width=0.8\linewidth]{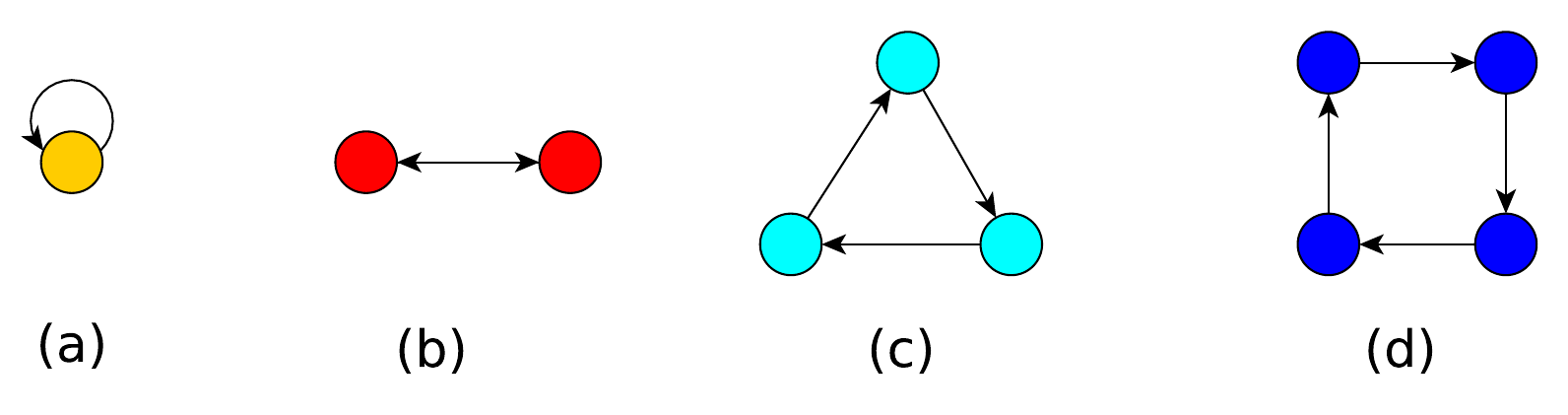}
    \caption{Types of loops in graphs: (a) self-loop, (b) loop with two nodes connected by an undirected edge, (c) and (d) are examples of n-node loops where n=3 and n=4 respectively.}
    \label{fig:loop}
\end{figure}

\subsection{Self-loops}
Sources of self-loops include:  
\begin{enumerate}  
    \item \textbf{Original Graph}: In some networks, such as webpage networks, a node (e.g., a webpage) might naturally link to itself.  
    \item \textbf{GNN Model Design}: Many GNN models, such as GCN \cite{kipfSemiSupervisedClassificationGraph2017} and DiG(ib) \cite{tongDirectedGraphConvolutional2020}, explicitly add self-loops to improve performance, particularly on homophilic graphs.  
\end{enumerate}

As discussed in Sec. \ref{lm:gcn_loop}, a \( k \)-layer GCN with self-loops would gather information from neighbors within the range of 0-hop to \( k \)-hop neighbors. This fact was established via matrix multiplication. In this section, we will prove it geometrically.
 
\begin{lemma}\label{lm:sloop}
    When self-loops are added to a graph, the \( k \)-hop neighbors of any node are also its \( (k+1) \)-hop neighbors.
\end{lemma}
The proof is provided in Appendix \ref{pf:sloop}.
This path-based property can be expressed in terms of the adjacency matrix:

\begin{lemma}
Let \( G \) be a graph with self-loops. Then for any \( k \geq 1 \), any connection present in \( A^k \) is also present in \( A^{k+1} \).
\end{lemma}

\subsection{Two-node Loops}
A directed graph where each pair of connected nodes has edges in both directions (making its adjacency matrix symmetric) can be viewed as an undirected graph. In other words, an undirected graph is equivalent to a directed graph where every edge is bidirected.

\begin{lemma} \label{lm:two_loop}
    For an undirected graph, for any \( k \geq 1 \), the \( k \)-hop neighbors of any node are also its \( (k+2) \)-hop neighbors.
\end{lemma}
The proof is provided in Appendix \ref{pf:two_loop}. This property can be expressed in terms of the adjacency matrix:

\begin{lemma}
Let \( G \) be an undirected graph. Then for any \( k \geq 1 \), any connection present in \( A^k \) is also present in \( A^{k+2} \).
\end{lemma}

\subsection{Multi-node Loops}
\begin{lemma}
\label{lm:mul_loop}
For a graph containing a loop of length \( m \), let \( v \) be any node in the graph. For any \( k \geq 1 \), if \( u \) is a \( k \)-hop neighbor of \( v \) where the \( k \)-hop path from \( u \) to \( v \) contains at least one node from the loop, then \( u \) is also a \( (k+m) \)-hop neighbor of \( v \).
\end{lemma}
The proof is provided in Appendix \ref{pf:mul_loop}. 
This path-based property can be naturally expressed in terms of the adjacency matrix of the graph:

\begin{lemma}
Let \( G \) be a graph containing a loop of length \( m \). Then for any \( k \geq 1 \), any connection present in \( A^k \) is also present in \( A^{k+m} \).
\end{lemma}

\subsection{Longest path}
\subsubsection{For Directed Graph}
\begin{lemma}
\label{ls:directed_largest_hop}
For a directed graph with adjacency matrix \( A \), if the graph contains no loops (cycles) and \( h \) is the length of the longest simple path, then:
\[ A^m = 0 \text{ for all } m > h \]
\end{lemma}
The proof is provided in Appendix \ref{pf:long_dire}.

\subsubsection{For Undirected Graph}
\begin{lemma}
\label{ls:undirected_largest_hop}
Let \( G \) be an undirected graph, and let \( h \) be the length of the longest path in \( G \). Then for any \( m > h \), the connections present in \( A^m \) are identical to those in \( A^h \).
\end{lemma}
The proof is provided in Appendix \ref{pf:long_undire}.

\subsection{Loops Influence}
Different types of graph structures influence how connectivity patterns evolve as we take higher powers of the adjacency matrix. Self-loops allow paths to extend by single steps, while two-node loops (undirected edges) enable extension by pairs of steps. More generally, any $m$-node loop allows paths to extend by $m$ steps while preserving all existing connections.

For the maximal path length, undirected and directed graphs behave quite differently. In undirected graphs, paths can always extend beyond the spanning tree's longest path length while maintaining the same connectivity pattern. However, in directed acyclic graphs, no paths can exist beyond this length.

Table~\ref{tab:graph_types} summarizes these relationships, where $\overset{c}{=}$ denotes identical connectivity patterns in the adjacency matrices.
\begin{table}[t]
\centering
\begin{tabular}{|l|l|}
\hline
\textbf{Graph Type} & \textbf{Matrix Connectivity} \\
\hline
Self-loops & $A^k \overset{c}{=} A^{k+1}$ \\
\hline  
Two-node loops & $A^k \overset{c}{=} A^{k+2}$ \\
\hline
$n$-node loops & $A^k \overset{c}{=} A^{k+n}$ \\
\hline
\hline
Undirected  & $A^m \overset{c}{=} A^h$ for $m > h$ \\
\hline
Directed acyclic & $A^m = 0$ for $m > h$ \\
\hline
\end{tabular}
\caption{Graph Types and Matrix Connectivity. Here $\overset{c}{=}$ denotes that two matrices have identical connectivity patterns (presence of non-zero entries), $h$ is the length of the longest path, and $m$ is any hop count value.}
\label{tab:graph_types}
\end{table}

We provide a mathematical analysis of how different neighborhood hops coexist in $k$-layer GNNs, a phenomenon that can contribute to over-smoothing. Our analysis identifies three critical factors that drive this coexistence:
(1) Adding self-loops; 
(2) Undirected edges;
(3) Presence of loops in datasets. 

\paragraph{Empirical Validation}
Common GNN preprocessing steps—adding self-loops and symmetrizing directed graphs through reverse edges—significantly increase the density of $k$-hop neighborhoods. Our experiments demonstrate that avoiding these modifications can prevent over-smoothing. 

\begin{figure}[ht]
\vskip 0.2in
    \centering

\captionsetup{font=small}
\begin{subfigure}{0.8\linewidth}
\includegraphics[width=\linewidth]{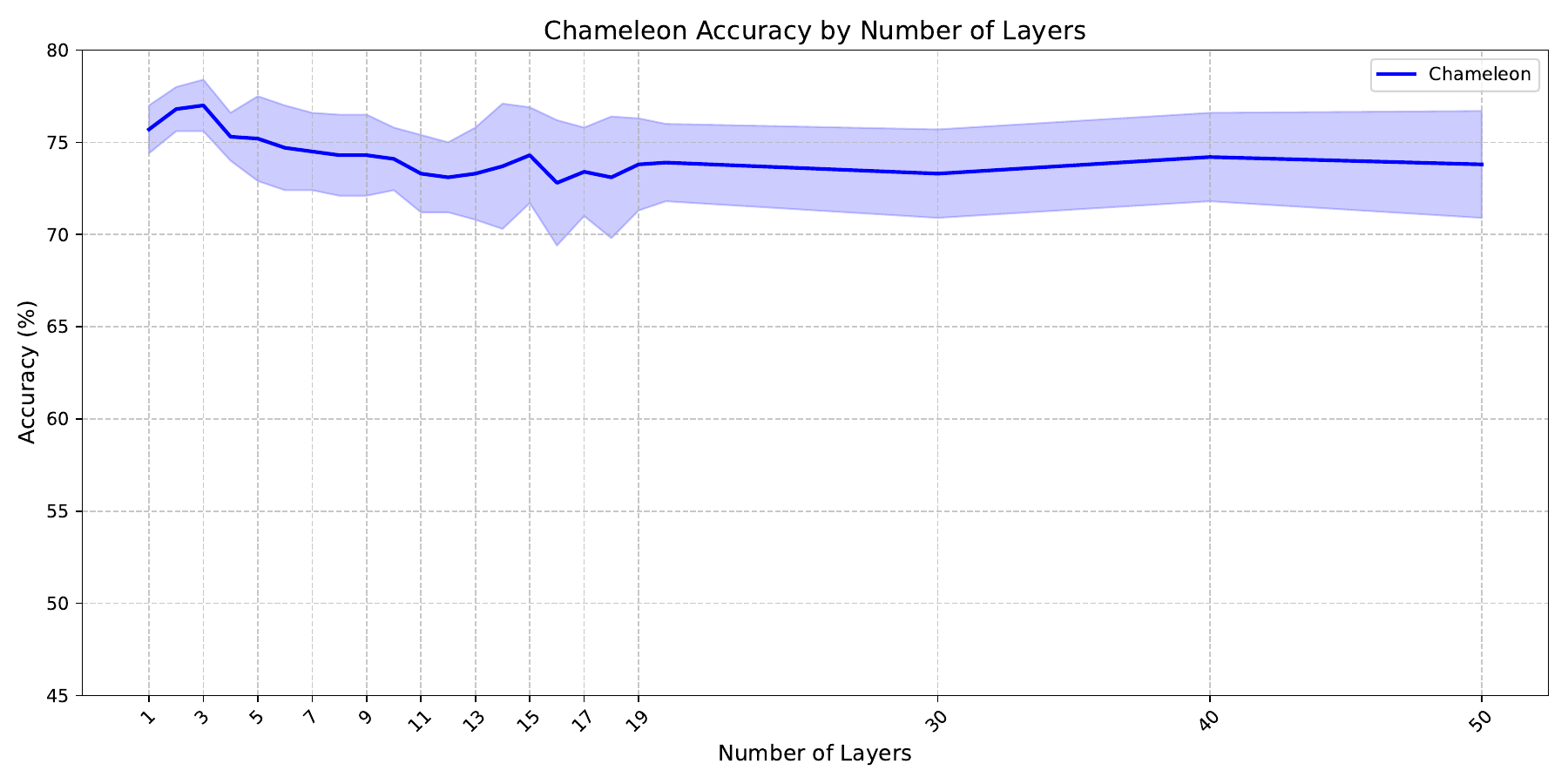}
\vskip -0.1in
\caption{Chameleon}
\label{fig_chameloon_50}
\end{subfigure}
\vskip 0.1in

\captionsetup{font=small}
\begin{subfigure}{0.8\linewidth}
\includegraphics[width=\linewidth]{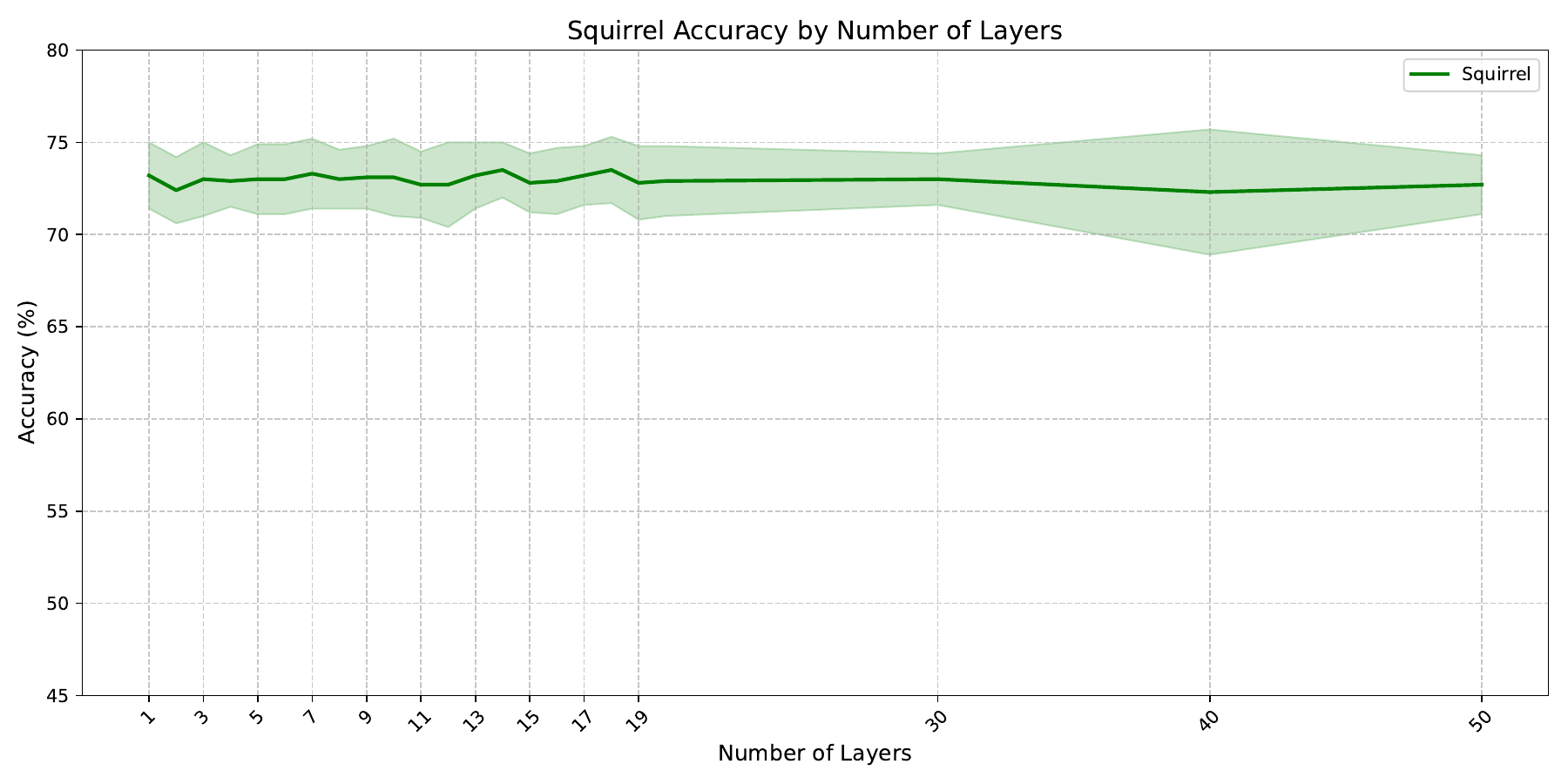}
\vskip -0.1in
\caption{Squirrel}
\label{fig_squirrel_50}
\end{subfigure}
\vskip -0.1in

\caption{Performance of Unidirectional GCN Without Self-loops on Chameleon and Squirrel Datasets: Model demonstrates stable accuracy up to 50 layers, with deeper architectures constrained by memory limitations. The solid line represents mean accuracy, while the shaded region indicates standard deviation across 10 data splits.}
    \label{fig:chame&squi_2hop}
    
\vskip -0.2in
\end{figure}

As shown in Fig.~\ref{fig:chame&squi_2hop}, a GCN model without self-loops or undirected conversion maintains performance even at 50 layers, whereas standard GNNs typically experience over-smoothing within 5 layers \cite{liDeeperInsightsGraph2018}. These results align with our theoretical analysis showing that self-loops and undirected edges induce neighborhood hop coexistence, a potential mechanism for over-smoothing.

More information about datasets and experiments are presented in Appendix \ref{ap:data}.

\section{Structure-Feature Dichotomy in Node Classification}
\label{sc:dichotomy}

Graph Neural Networks (GNNs) combine node features and graph structure for predictions. However, recent work shows structure-agnostic models like MLPs outperform GNNs on certain datasets (e.g., WebKB \cite{zhengColdBrewDistilling2022}). Complementing this finding, we show that some node classification tasks perform equally well without node features. Based on this Structure-Feature Dichotomy, we categorize tasks into three types: feature-only, structure-only, and hybrid. We then analyze how GNNs make predictions for the latter two cases.

\subsection{Structure-feature Hybrid Type}
Citation networks like Cora, CiteSeer, and PubMed represent classic node classification tasks where research papers are categorized by their topics. While individual papers may not contain comprehensive field-specific content, aggregating features from neighboring nodes can enrich the representation of each paper's research domain, making GNNs particularly effective for this task.

\begin{figure}
\centering
\captionsetup{font=small}
\begin{subfigure}{0.7\linewidth}
\includegraphics[width=\linewidth]{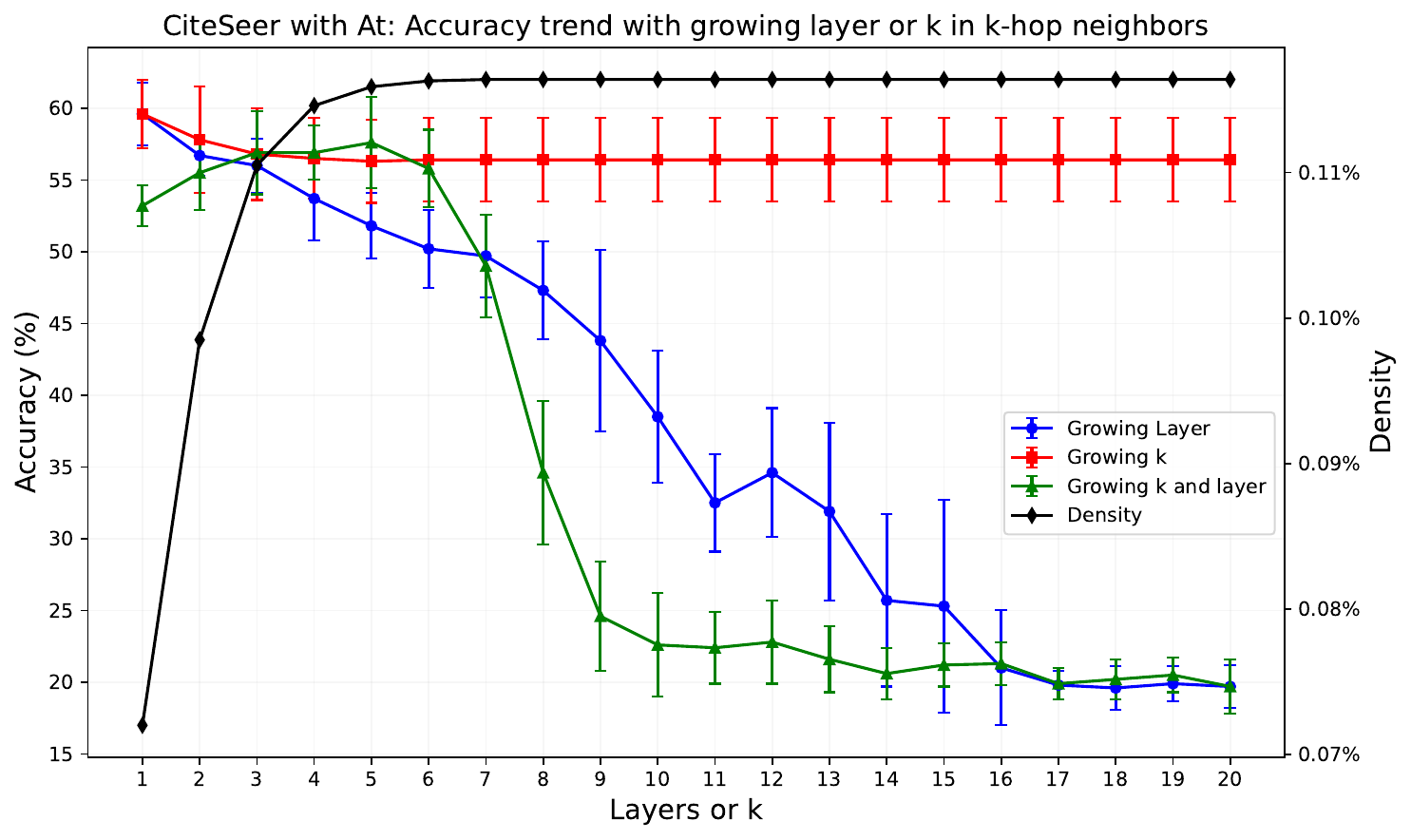}
\vskip -0.1in
\caption{CiteSeer}
\label{fig_cite_A}
\end{subfigure}
\vskip 0.1in

\begin{subfigure}{0.7\linewidth}
\includegraphics[width=\linewidth]{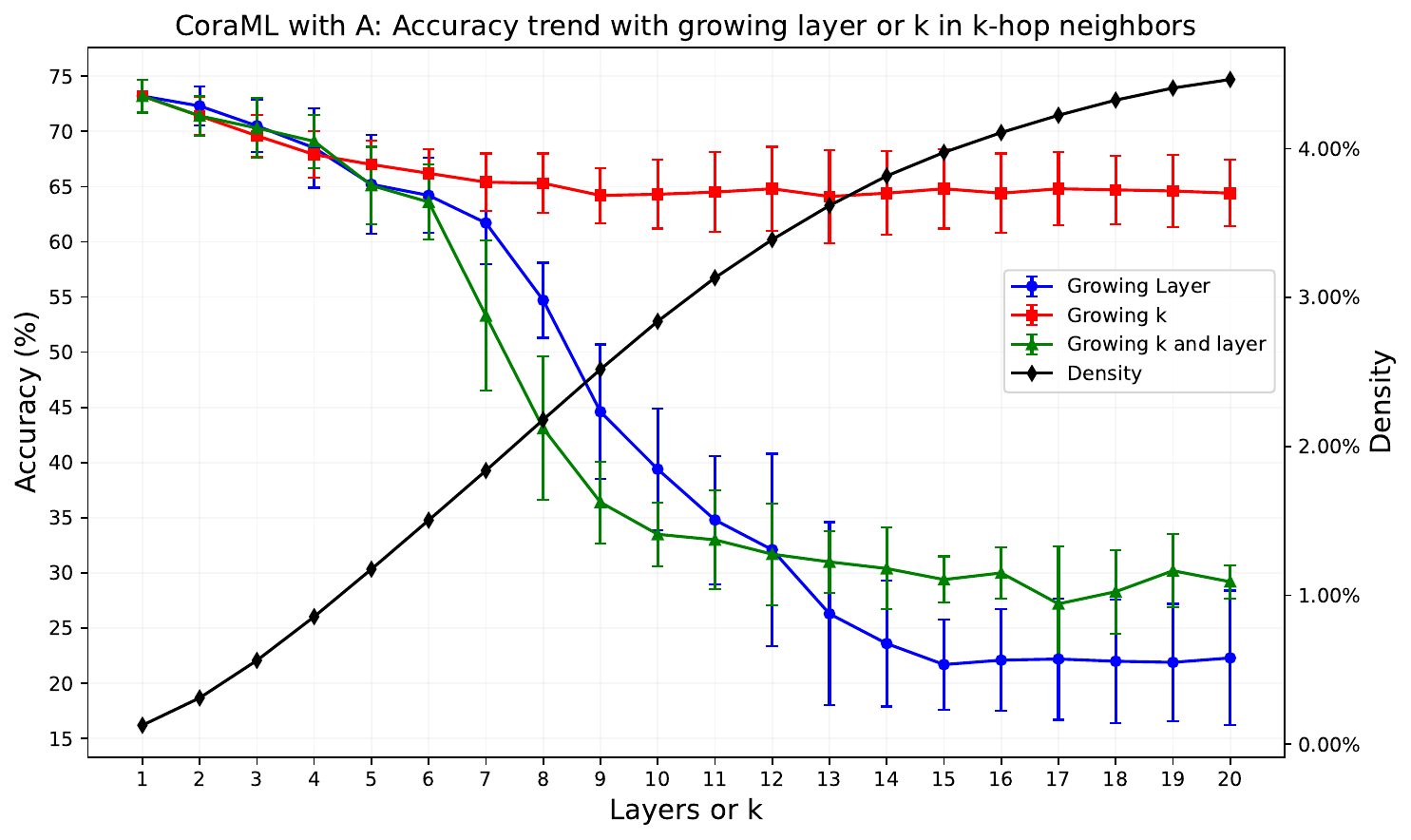}
\vskip -0.1in
\caption{CoraML}
\label{fig_cora_A}
\end{subfigure}
\vskip 0.1in

\begin{subfigure}{0.7\linewidth}
\includegraphics[width=\linewidth]{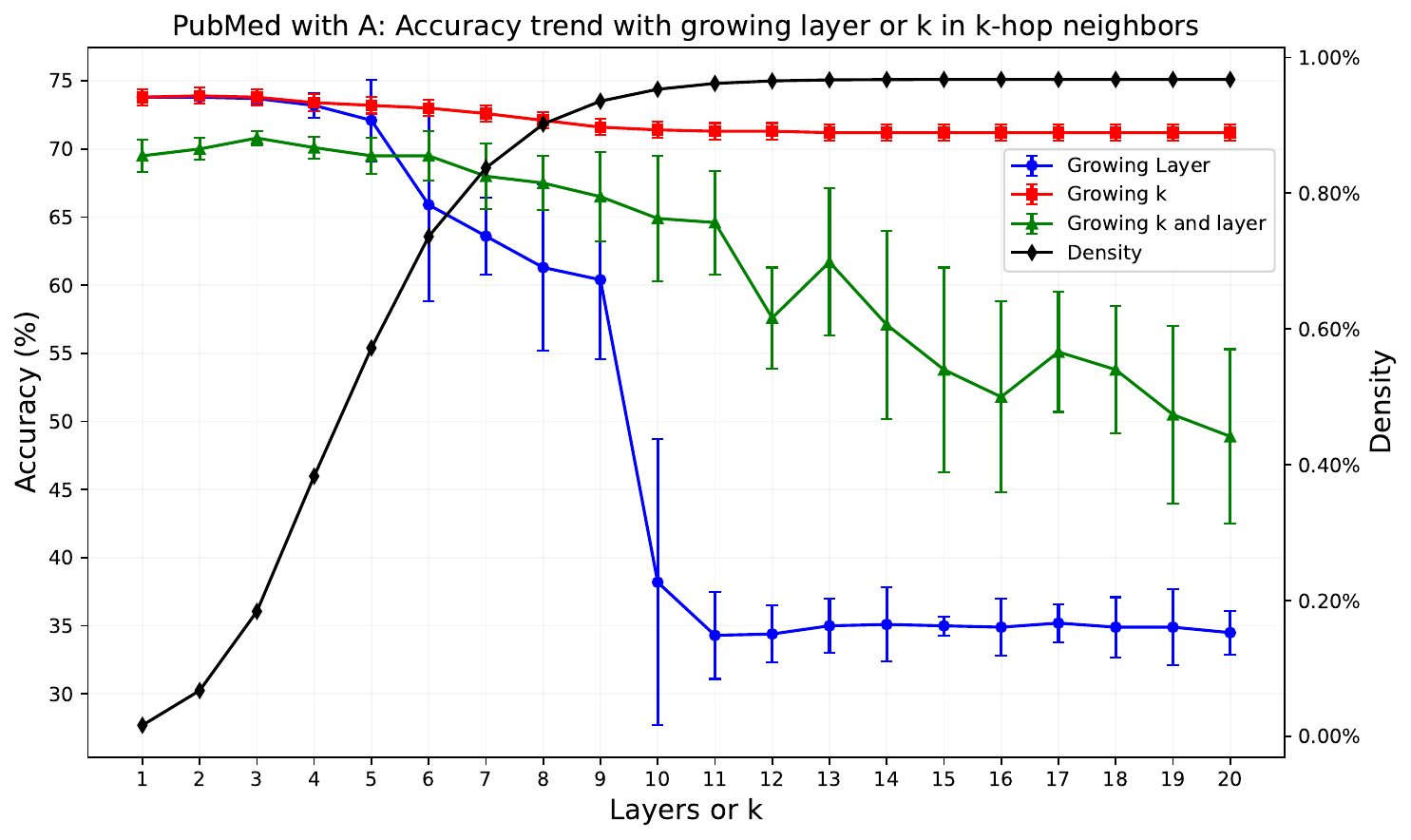}
\vskip -0.1in
\caption{PubMed}
\label{fig_pub_A}
\end{subfigure}

\vskip -0.15in
\caption{Comparison of different GCN architectures on three datasets: $k$-layer GCN (\textbf{blue}), $1$-layer GCN with $k$-hop neighbors (\textbf{red}), and $k$-hop neighbors with $1$-layer GCN and ($k$-1) linear layers (\textbf{green}). The black line shows the density of $k$-hop adjacency matrix.}

\label{fig:A_all}
\end{figure}
Figure \ref{fig:A_all} shows consistent patterns across CiteSeer, CoraML, and PubMed datasets comparing three approaches: (1) increasing GNN layers with first-order neighbors (blue line), (2) single-layer GNN with increasing $k$-hop neighbors (red line), and (3) $k$-hop neighbors with ($k$-1) additional linear layers (green line). The black line represents the density of the effective adjacency matrix---the percentage of non-zero elements after $k$-hop expansion. The low density values across all datasets indicate these are sparse directed networks. 

The single-layer GNN with increasing $k$-hop neighbors maintains stable performance, while both the deep GNN and the hybrid approach show significant performance degradation with increasing depth.
While both architectures access $k$-hop neighborhood information---through $A^k$ in single-layer GNN and k successive applications of $A$ in $k$-layer GNN---their empirical performance differs substantially despite theoretical equivalence in terms of Universal Approximation (Section \ref{sc:sum_k_uat}).

The fact that performance remains stable when increasing the neighborhood size k in a single-layer architecture (red line) indicates minimal over-smoothing in this case. Thus, we hypothesized that gradient-related issues might be the primary cause of performance degradation in deeper networks.

To test this hypothesis, we designed approach (3) which combines $k$-hop neighborhood aggregation with ($k$-1) additional linear layers. This architecture shares parameter count with the deep GNN while using expanded neighborhoods like the single-layer approach. The deteriorating performance of this hybrid approach parallels that of the deep GNN, strongly suggesting that gradient-related issues, rather than over-smoothing, cause the performance drop in deep GNNs.

\begin{figure}
\centering
\captionsetup[subfigure]{font=scriptsize}
\begin{subfigure}{0.7\linewidth}
\includegraphics[width=\linewidth]{figs/CiteSeer+At.pdf}
\vskip -0.1in
\caption{$A^T$ as adjacency matrix}
\label{fig_cite_At}
\end{subfigure}
\vskip 0.1in
\begin{subfigure}{0.7\linewidth}
\includegraphics[width=\linewidth]{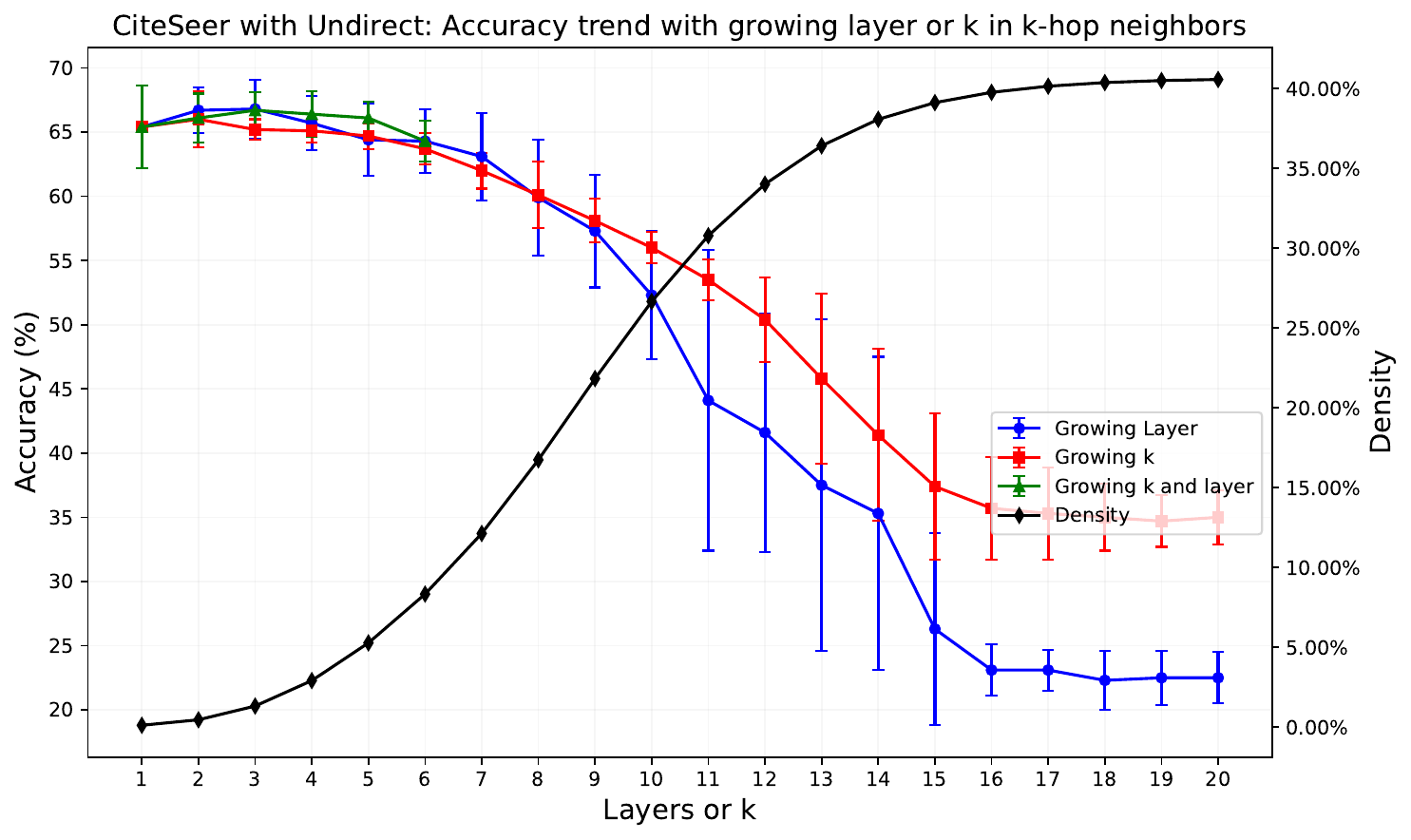}
\vskip -0.1in
\caption{Bidirectional Propagation}
\label{fig_cite_undirect}
\end{subfigure}

\vskip -0.15in


\caption{Comparison of different GCN architectures on CiteSeer dataset under different adjacency matrix formulations. (Top) Using transposed adjacency matrix $A^T$, which propagates information from cited papers to citing papers. (Bottom) Using undirected graph adjacency matrix $A + A^T$, which enables bidirectional information flow. In each subplot: k-layer GCN (blue), 1-layer GCN with k-hop neighbors (red), and k-hop neighbors with (k-1) linear layers (green). The black line indicates the density of the k-hop adjacency matrix.}

\label{fig:cite_2}
\end{figure}
We further investigated Reverse Direction and Bidirectional Propagation, with experimental results on CiteSeer presented in Figure \ref{fig:cite_2}. For Reverse Direction Propagation, we observed performance trends similar to forward direction propagation. However, Bidirectional Propagation exhibited distinct behavior: increasing the neighborhood size k in single-layer architectures (red line) led to performance degradation, which can be attributed to over-smoothing since the connection density inversely correlates with k. The connection density (black line) and performance with increasing k (red line) both stabilized after k=17. 

As shown in Figure \ref{fig_cite_undirect}, $k$-layer GNNs with first-order neighbors (blue line) performed worse than single-layer models with equivalent $k$-hop neighbors (red line), likely due to the compound effects of over-smoothing and vanishing gradients.

Experimental results for the CoraML and PubMed datasets are detailed in Appendix \ref{ap_add_experi}. While Reverse Direction Propagation exhibits performance trends similar to our main findings, Bidirectional Propagation demonstrates significant over-smoothing behavior. Specifically, in the bidirectional case, neighborhood density exceeds 80\% within 8 hops, leading to degraded performance. Consequently, single-layer models utilizing equivalent $k$-hop neighborhoods underperform compared to $k$-layer GNNs that only aggregate first-order neighbors.

Thus, while our findings hold true for sparse networks, dense networks exhibit different behavior. The distinct performance patterns in dense networks suggest that over-smoothing may still play a significant role in their degradation.

These results enhance our understanding of GNN performance degradation, indicating that optimizing architectural design for effective gradient flow may be more crucial than addressing over-smoothing effects in sparse networks, while over-smoothing remains a key consideration for dense networks.

\subsection{Structure-only Type}

In this section, we will present three datasets which work well without node features, where all nodes have uniform features.

\begin{table}[htbp]

    \captionsetup{font=small} 
\caption{Classification accuracy (\%) of Dir-GNN \cite{rossiEdgeDirectionalityImproves2024} with different feature configurations and normalization schemes on Chameleon, Squirrel and Telegram datasets. Feature configurations include: original node features from datasets (Origin Feature), constant features (No Feature, all set to 1), and node degree variants (in-degree, out-degree, or both). \textbf{Bold} values indicate learning failure with row normalization and no features. \underline{Underlined} values show the worst performance among all configurations, except for the case of row normalization with no features, which can be attributed to numerical instability when normalization is absent.}
\label{tb:all1_chame_squ}
    \centering
\setlength{\tabcolsep}{1.5pt}
    \vskip 0.15in
\begin{center}
\begin{small}
    \begin{tabular}{lccccc}
    \toprule
      \textbf{Chameleon} & \textbf{Feature} &\textbf{None} &  \textbf{Row} & \textbf{Sym} & \textbf{Dir}  \\
      \midrule
      MLP &  \multicolumn{3}{c}{}48.0±1.6   &\\
      \midrule
Origin Feature & 2,325 & \underline{79.1±1.4} & 80.0±1.5 & 79.4±1.6 & 79.8±1.4 \\
       No Feature & 1 & \underline{73.6±2.4} & \textbf{23.0±2.6} & 77.9±1.8 & 78.1±1.2\\
In-degree & 1 &\underline{75.2±2.0} &  77.0±2.0 & 78.0±2.5& 78.1±1.8  \\
Out-degree & 1 & \underline{73.6±2.0} &  77.8±1.5 & 78.0±2.3 & 77.6±2.1  \\
       Both degrees & 2 &\underline{75.5±1.8} &  77.6±1.5 & 77.9±2.2 & 77.4±1.3  \\
       \midrule
       \midrule
\textbf{Squirrel} & \textbf{Feature} & \textbf{None} &  \textbf{Row} & \textbf{Sym} & \textbf{Dir}  \\
      \midrule
      MLP &  \multicolumn{3}{c}{}36.3±1.5  \\
      \midrule
       Origin Feature& 2,089 & \underline{74.3±2.3} & 75.1±1.6 & 76.3±1.9 & 76.1±2.0\\
       No Feature& 1 & \underline{67.8±3.4}  & \textbf{19.5±1.1} & 75.5±2.3 & 75.6±2.0\\
    In-degree & 1 & \underline{64.9±5.8} &  73.0±3.0 
    & 75.7±1.7 & 75.3±1.6  \\
Out-degree & 1 & \underline{63.7±4.9} &  72.2±3.8 & 75.5±1.6 & 75.1±1.8  \\
Both degrees & 2 & \underline{67.9±3.9} &  73.1±3.1 & 76.1±1.4 & 75.4±1.8  \\
       \midrule
       \midrule
\textbf{Telegram} & \textbf{Feature} & \textbf{None} &  \textbf{Row} & \textbf{Sym} & \textbf{Dir}  \\
      \midrule
 Origin Feature& 1 & 95.6±2.8&	74.2±5.5	&93.0±4.1&	92.8±4.7\\
 No Feature& 1 & 95.4±4.0	& \textbf{38.0±0.0}&	93.0±4.7&	93.0±3.0 \\
       \bottomrule
    \end{tabular}
\end{small}
\end{center}
\vskip -0.1in
    
    \label{tab:all1}
\end{table}
As shown in Table \ref{tb:all1_chame_squ}, for Chameleon, Squirrel and Telegram datasets, Dir-GNN  \footnote{Specific parameter settings are shown in ScaleNet \cite{jiangScaleInvarianceGraph2024} . For Telegram datasets, self-loops are added for better performance.} \cite{rossiEdgeDirectionalityImproves2024} predicts as well with no feature as with original features. We will give a brief summarization of existing normalizations.

\subsubsection{Normalizations}
Graph normalization, which typically involves dot multiplication of the adjacency matrix to adjust edge weights, plays a crucial role in graph neural networks (GNNs). While various normalization schemes exist, their theoretical implications remain under-explored. We denote a general normalization function as $f(A)$.

\paragraph{No Normalization} The simplest approach is to use the raw adjacency matrix without any normalization \cite{liGatedGraphSequence2017}:
$f_1(A) = A$.
In this case, the node feature update rule becomes:
\[
h_i^{(l+1)} = \sigma(\sum_{j \in \mathcal{N}(i)} h_j^{(l)}W^{(l)})
\]
The aggregation directly sums neighboring features, leading to larger feature magnitudes for higher degree nodes. With homogeneous features $h_i^{(0)} = 1$, node representations become proportional to degrees.

While this makes it suitable for degree-dependent tasks like traffic prediction and network flow classification, repeated aggregation of unnormalized features can cause numerical instability. The node representations may grow or vanish exponentially with network depth. This numerical instability explains the suboptimal performance of unnormalized adjacency matrices compared to normalized variants in Table \ref{tb:all1_chame_squ}. The exponential growth or decay of node representations across layers likely hindered the model's ability to learn effective graph representations, despite preserving the degree information.

\paragraph{Row Normalization} Row normalization \cite{hamiltonInductiveRepresentationLearning2018} scales each row of the adjacency matrix by the inverse of node degree:
$f_2(A) = D^{-1}A$. 
The node feature update rule becomes:
\[
h_i^{(l+1)} = \sigma(\frac{\sum_{j \in \mathcal{N}(i)} h_j^{(l)}W^{(l)}}{d_i})
\]
For this formulation, the aggregated information represents the mean of neighboring features rather than their sum. Node degrees no longer directly influence feature magnitudes. With homogeneous features $h_i^{(0)} = 1$, all nodes get identical representations.
This explains the poor traffic prediction in Table \ref{tb:all1_chame_squ}---degree information is lost.

\paragraph{Symmetric Normalization} Symmetric normalization \cite{kipfSemiSupervisedClassificationGraph2017} applies:
$f_3(A) = D^{-1/2}AD^{-1/2}$. 
The node feature update rule becomes:
\[
h_i^{(l+1)} = \sigma(\sum_{j \in \mathcal{N}(i)} \frac{h_j^{(l)}W^{(l)}}{\sqrt{d_i d_j}} )
\]
The neighbor's influence is determined by both degrees---if a neighbor's degree is much larger than the center node's, its feature weight becomes smaller than in row normalization.
\paragraph{Directed Normalization} For directed graphs, \citet{rossiEdgeDirectionalityImproves2024} proposes:
$f_4(A) = D_{in}^{-1/2}AD_{out}^{-1/2}$. The node feature update rule becomes:
\[
h_i^{(l+1)} = \sigma(\sum_{j \in \mathcal{N}(i)} \frac{h_j^{(l)}W^{(l)}}{\sqrt{d_i^{in}d_j^{out}}} )
\]
This distinguishes between in-degree and out-degree for more accurate normalization in directed graphs.

In summary, row normalization of adjacency matrices, when applied with uniform node features, results in loss of structural information since all nodes become indistinguishable. While using the unnormalized adjacency matrix preserves both degree information and feature distinctions, it can lead to numerical instability since the eigenvalues of $f(A)$ may grow or diminish exponentially, rather than being bounded within [-1, 1]. This instability can affect the training of graph neural networks.

\subsubsection{How GNN Predict for Structure-only Classification}

\begin{figure}
    \captionsetup{font=small}
\vskip 0.2in
    \centering
    \includegraphics[width=0.8\linewidth]{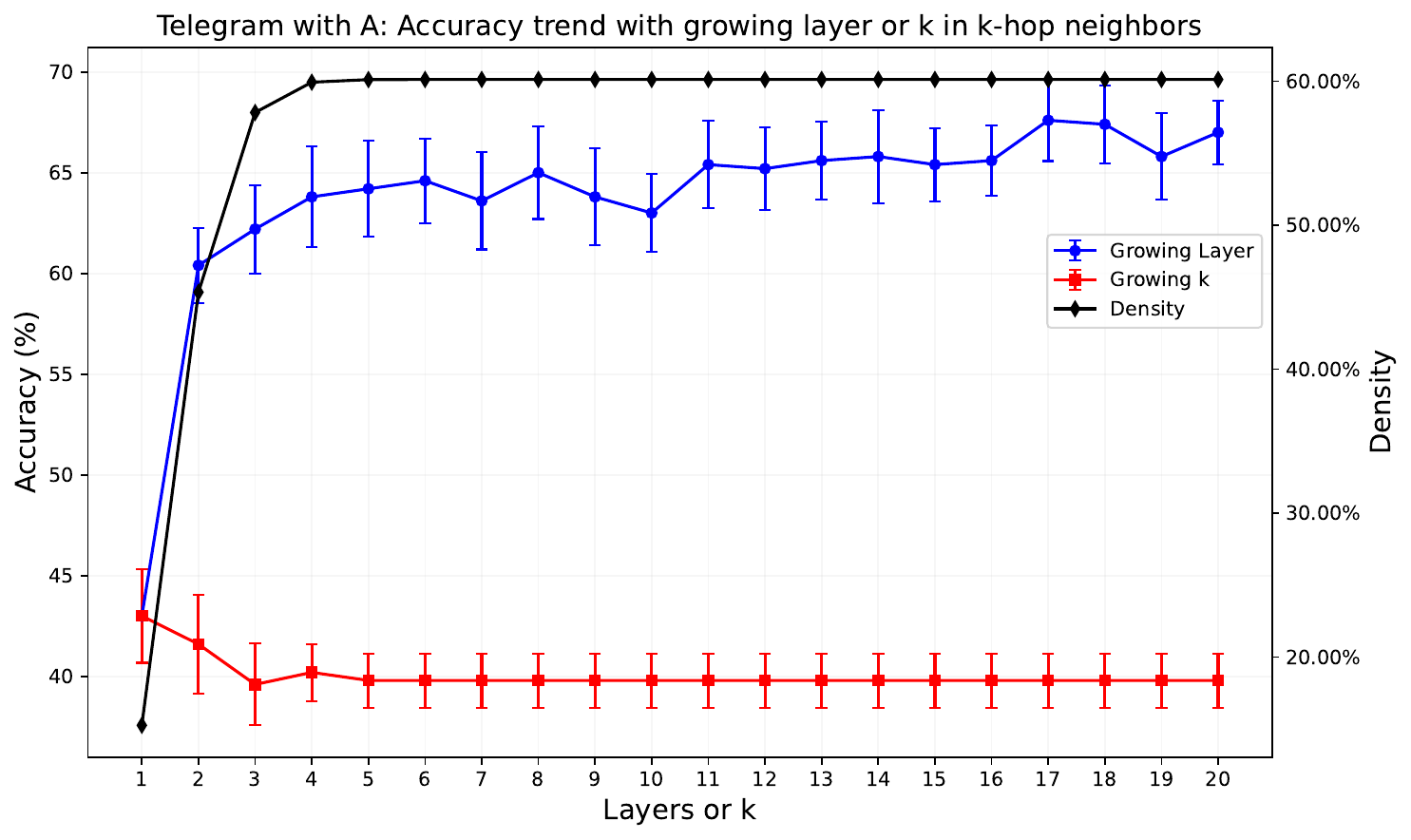}
    \vskip -0.1in
    \caption{On Telegram dataset, layer growth achieves good accuracy while k growth performs poorly with no predictive power. Density shows the percentage of non-zero elements in the equivalent adjacency matrix.}
    \label{fig:tel_3}
\end{figure}
Fig. \ref{fig:tel_3} demonstrates that increasing the number of GCN layers leads to better accuracy, while increasing the power $k$ of the adjacency matrix ($k$-hop neighborhoods) results in poor predictive performance on the Telegram dataset. 

As shown in Table \ref{tb:all1_chame_squ}, models using uniform features (all set to 1) achieve comparable performance to those using original features, suggesting that the original node features contribute little to predictive power. We therefore focus our analysis on the simpler uniform feature case.

With uniform features set to 1, we observe two scenarios:
\begin{itemize}
    \item Single-layer GCN with $k$-hop adjacency matrix ($A^k$): The node representations simply become counts of $k$-hop neighbors. The model loses predictive power beyond $k>2$.
    \item $k$-layer GCN with standard adjacency matrix ($A$): After the first layer, node representations become equivalent to node degrees. In subsequent layers, representations capture the degrees of neighboring nodes. This effectively means a ($k+1$)-layer GNN without features is equivalent to a $k$-layer GNN using node degrees as the only feature.
\end{itemize}

This explains the patterns observed in Figure \ref{fig:tel_3}. The single-layer GNN performs poorly because it only considers individual node degrees without incorporating neighborhood information. From two layers onward, performance improves steadily as the effective adjacency matrix becomes denser, eventually stabilizing when additional layers no longer increase the density of connections.

\subsection{Summary}

In this section, we further analyze MPNN predictions through the structure-feature dichotomy.
Node classification datasets can be divided into a structure-feature dichotomy, where some datasets perform better with MLPs. For datasets where GNNs are effective, predictions can be made with or without node features, determined by the task type.

\paragraph{Structure-feature Hybrid Tasks}
Content-based classification problems, exemplified by citation networks, require integration of both node features and neighborhood features propagated through structural connections. Our analysis reveals that the commonly observed performance degradation in deeper layers, traditionally attributed to over-smoothing, may instead stem from gradient-related challenges, especially for sparse networks.

\paragraph{Structure-only tasks}
In tasks such as traffic prediction and network flow classification, MPNNs can make predictions using purely structural information. We show that when nodes have uniform features, the prediction mechanism shifts---node degree becomes the effective feature, and a (k+1)-layer MPNN with uniform features predicts equivalently to a $k$-layer MPNN using node degree as the sole feature.

We also prove that combining featureless inputs with row normalization leads to degenerate predictions where MPNNs learn nothing.

In sum, structure-only tasks can be considered as a special type of Structure-feature Hybrid Tasks, where the node degree act as node feature.

\section{Conclusions}

This work demystifies Message Passing Neural Networks by revealing their computational essence: the message passing process is fundamentally a memory-efficient implementation of matrix multiplication operations. We establish that $k$-layer MPNNs aggregate information from $k$-hop neighborhoods through iterative computation, making them practical for large graphs where direct computation of powered adjacency matrices would be prohibitively expensive.

Through careful analysis of loop structures, we theoretically characterize how different types of loops influence $k$-hop neighborhood density. We demonstrate that common GNN practices, such as adding self-loops and converting directed graphs to undirected ones by adding reverse edges, significantly increase $k$-hop neighborhood density, potentially leading to over-smoothing.

Our analysis challenges two common misconceptions in the field: (1) performance degradation in deeper GNNs is not necessarily due to over-smoothing. For sparse directed graphs, deeper architectures are less susceptible to over-smoothing due to low connection density, yet their performance degrades due to vanishing gradients and overfitting from accumulated weights; and (2) deeper GNN architectures do not necessarily lead to over-smoothing as long as loop structures don't create dense $k$-hop connectivity.

Furthermore, we explained how GNNs work in structure-feature hybrid tasks and how for structure-only tasks, the node degree becomes the actual feature.

These insights offer theoretical understanding of how GNNs work and provides practical guidance for GNN architecture design, particularly regarding the choice of directed versus undirected aggregation, whether to add self-loops, and the selection of normalization strategies.


\bibliography{demy_mpnn}

\begin{thebibliography}{24}
\providecommand{\natexlab}[1]{#1}
\providecommand{\url}[1]{\texttt{#1}}
\expandafter\ifx\csname urlstyle\endcsname\relax
  \providecommand{\doi}[1]{doi: #1}\else
  \providecommand{\doi}{doi: \begingroup \urlstyle{rm}\Url}\fi

\bibitem[Bechler-Speicher et~al.(2024)Bechler-Speicher, Amos, Gilad-Bachrach,
  and Globerson]{bechler-speicherGraphNeuralNetworks2024}
Bechler-Speicher, M., Amos, I., Gilad-Bachrach, R., and Globerson, A.
\newblock Graph {Neural} {Networks} {Use} {Graphs} {When} {They} {Shouldn}'t.
\newblock June 2024.

\bibitem[Chamberlain et~al.(2021)Chamberlain, Rowbottom, Gorinova, Bronstein,
  Webb, and Rossi]{chamberlainGRANDGraphNeural2021a}
Chamberlain, B., Rowbottom, J., Gorinova, M.~I., Bronstein, M., Webb, S., and
  Rossi, E.
\newblock {GRAND}: {Graph} {Neural} {Diffusion}.
\newblock In \emph{Proceedings of the 38th {International} {Conference} on
  {Machine} {Learning}}, pp.\  1407--1418. PMLR, July 2021.
\newblock ISSN: 2640-3498.

\bibitem[Dehmamy et~al.(2019)Dehmamy, Barabasi, and
  Yu]{dehmamyUnderstandingRepresentationPower2019}
Dehmamy, N., Barabasi, A.-L., and Yu, R.
\newblock Understanding the {Representation} {Power} of {Graph} {Neural}
  {Networks} in {Learning} {Graph} {Topology}.
\newblock In \emph{Advances in {Neural} {Information} {Processing} {Systems}},
  volume~32. Curran Associates, Inc., 2019.

\bibitem[Gilmer et~al.(2017)Gilmer, Schoenholz, Riley, Vinyals, and
  Dahl]{gilmerNeuralMessagePassing2017}
Gilmer, J., Schoenholz, S.~S., Riley, P.~F., Vinyals, O., and Dahl, G.~E.
\newblock Neural {Message} {Passing} for {Quantum} {Chemistry}.
\newblock In \emph{Proceedings of the 34th {International} {Conference} on
  {Machine} {Learning}}, pp.\  1263--1272. PMLR, July 2017.
\newblock ISSN: 2640-3498.

\bibitem[Hamilton et~al.(2017)Hamilton, Ying, and
  Leskovec]{hamiltonInductiveRepresentationLearning2018}
Hamilton, W., Ying, Z., and Leskovec, J.
\newblock Inductive representation learning on large graphs.
\newblock \emph{Advances in neural information processing systems}, 30, 2017.

\bibitem[Hornik()]{hornikApproximationCapabilitiesMuitilayer}
Hornik, K.
\newblock Approximation {Capabilities} of {Muitilayer} {Feedforward}
  {Networks}.

\bibitem[Hornik et~al.(1989)Hornik, Stinchcombe, and White]{HornikKurt1989Mfna}
Hornik, K., Stinchcombe, M., and White, H.
\newblock Multilayer feedforward networks are universal approximators.
\newblock \emph{Neural networks}, 2\penalty0 (5):\penalty0 359--366, 1989.
\newblock ISSN 0893-6080.

\bibitem[Jiang et~al.(2024)Jiang, Wang, Lones, and
  Pang]{jiangScaleInvarianceGraph2024}
Jiang, Q., Wang, C., Lones, M., and Pang, W.
\newblock Scale {Invariance} of {Graph} {Neural} {Networks}, December 2024.
\newblock arXiv:2411.19392 [cs].

\bibitem[Kipf \& Welling(2016)Kipf and
  Welling]{kipfSemiSupervisedClassificationGraph2017}
Kipf, T.~N. and Welling, M.
\newblock Semi-supervised classification with graph convolutional networks.
\newblock \emph{arXiv preprint arXiv:1609.02907}, 2016.

\bibitem[Li et~al.(2018)Li, Han, and Wu]{liDeeperInsightsGraph2018}
Li, Q., Han, Z., and Wu, X.-m.
\newblock Deeper {Insights} {Into} {Graph} {Convolutional} {Networks} for
  {Semi}-{Supervised} {Learning}.
\newblock \emph{Proceedings of the AAAI Conference on Artificial Intelligence},
  32\penalty0 (1), April 2018.
\newblock ISSN 2374-3468.
\newblock \doi{10.1609/aaai.v32i1.11604}.
\newblock Number: 1.

\bibitem[Li et~al.(2017)Li, Tarlow, Brockschmidt, and
  Zemel]{liGatedGraphSequence2017}
Li, Y., Tarlow, D., Brockschmidt, M., and Zemel, R.
\newblock Gated {Graph} {Sequence} {Neural} {Networks}, September 2017.
\newblock URL \url{http://arxiv.org/abs/1511.05493}.
\newblock arXiv:1511.05493 [cs].

\bibitem[Maurya et~al.(2022)Maurya, Liu, and
  Murata]{mauryaSimplifyingApproachNode2022}
Maurya, S.~K., Liu, X., and Murata, T.
\newblock Simplifying approach to node classification in {Graph} {Neural}
  {Networks}.
\newblock \emph{Journal of Computational Science}, 62:\penalty0 101695, July
  2022.
\newblock ISSN 1877-7503.
\newblock \doi{10.1016/j.jocs.2022.101695}.

\bibitem[Park et~al.(2024)Park, Heo, and
  Kim]{parkMitigatingOversmoothingReverse2024}
Park, M., Heo, J., and Kim, D.
\newblock Mitigating {Oversmoothing} {Through} {Reverse} {Process} of {GNNs}
  for {Heterophilic} {Graphs}.
\newblock In \emph{Proceedings of the 41st {International} {Conference} on
  {Machine} {Learning}}, June 2024.

\bibitem[Pei et~al.(2020)Pei, Wei, Chang, Lei, and
  Yang]{peiGeomGCNGeometricGraph2020}
Pei, H., Wei, B., Chang, K. C.-C., Lei, Y., and Yang, B.
\newblock Geom-gcn: Geometric graph convolutional networks.
\newblock \emph{arXiv preprint arXiv:2002.05287}, 2020.

\bibitem[Rossi et~al.(2024)Rossi, Charpentier, Di~Giovanni, Frasca, Günnemann,
  and Bronstein]{rossiEdgeDirectionalityImproves2024}
Rossi, E., Charpentier, B., Di~Giovanni, F., Frasca, F., Günnemann, S., and
  Bronstein, M.~M.
\newblock Edge directionality improves learning on heterophilic graphs.
\newblock In \emph{Learning on {Graphs} {Conference}}, pp.\  25--1. PMLR, 2024.

\bibitem[Rozemberczki et~al.(2021)Rozemberczki, Allen, and
  Sarkar]{rozemberczki2021multi}
Rozemberczki, B., Allen, C., and Sarkar, R.
\newblock Multi-scale attributed node embedding.
\newblock \emph{Journal of Complex Networks}, 9\penalty0 (2):\penalty0 cnab014,
  2021.

\bibitem[Rusch et~al.(2023)Rusch, Bronstein, and
  Mishra]{ruschSurveyOversmoothingGraph2023}
Rusch, T.~K., Bronstein, M.~M., and Mishra, S.
\newblock A {Survey} on {Oversmoothing} in {Graph} {Neural} {Networks}, March
  2023.
\newblock arXiv:2303.10993 [cs].

\bibitem[Stanovic et~al.(2025)Stanovic, Gaüzère, and
  Brun]{stanovicGraphNeuralNetworks2025}
Stanovic, S., Gaüzère, B., and Brun, L.
\newblock Graph {Neural} {Networks} with maximal independent set-based pooling:
  {Mitigating} over-smoothing and over-squashing.
\newblock \emph{Pattern Recognition Letters}, 187:\penalty0 14--20, January
  2025.
\newblock ISSN 01678655.
\newblock \doi{10.1016/j.patrec.2024.11.004}.

\bibitem[Tong et~al.(2020)Tong, Liang, Sun, Rosenblum, and
  Lim]{tongDirectedGraphConvolutional2020}
Tong, Z., Liang, Y., Sun, C., Rosenblum, D.~S., and Lim, A.
\newblock Directed graph convolutional network.
\newblock \emph{arXiv preprint arXiv:2004.13970}, 2020.

\bibitem[Xie et~al.(2020)Xie, Li, Yang, Wong, and Han]{xieWhenGNNsWork2020}
Xie, Y., Li, S., Yang, C., Wong, R. C.-W., and Han, J.
\newblock When {Do} {GNNs} {Work}: {Understanding} and {Improving}
  {Neighborhood} {Aggregation}.
\newblock \emph{IJCAI'20: Proceedings of the Twenty-Ninth International Joint
  Conference on Artificial Intelligence, \{IJCAI\} 2020}, 2020\penalty0 (1),
  July 2020.
\newblock \doi{10.24963/ijcai.2020/181}.

\bibitem[Xu et~al.(2019)Xu, Hu, Leskovec, and Jegelka]{xuHowPowerfulAre2019}
Xu, K., Hu, W., Leskovec, J., and Jegelka, S.
\newblock How {Powerful} are {Graph} {Neural} {Networks}?, February 2019.
\newblock arXiv:1810.00826 [cs].

\bibitem[Zhang et~al.(2021)Zhang, He, Brugnone, Perlmutter, and
  Hirn]{zhangMagNetNeuralNetwork}
Zhang, X., He, Y., Brugnone, N., Perlmutter, M., and Hirn, M.
\newblock Magnet: A neural network for directed graphs.
\newblock \emph{Advances in neural information processing systems},
  34:\penalty0 27003--27015, 2021.

\bibitem[Zheng et~al.(2022)Zheng, Huang, Rao, Katariya, Wang, and
  Subbian]{zhengColdBrewDistilling2022}
Zheng, W., Huang, E.~W., Rao, N., Katariya, S., Wang, Z., and Subbian, K.
\newblock Cold {Brew}: {Distilling} {Graph} {Node} {Representations} with
  {Incomplete} or {Missing} {Neighborhoods}, March 2022.
\newblock URL \url{http://arxiv.org/abs/2111.04840}.
\newblock arXiv:2111.04840 [cs].

\bibitem[Zhu et~al.(2020)Zhu, Yan, Zhao, Heimann, Akoglu, and
  Koutra]{zhu2020beyond}
Zhu, J., Yan, Y., Zhao, L., Heimann, M., Akoglu, L., and Koutra, D.
\newblock Beyond homophily in graph neural networks: Current limitations and
  effective designs.
\newblock \emph{Advances in neural information processing systems},
  33:\penalty0 7793--7804, 2020.

\end{thebibliography}
\bibliographystyle{icml2025}

\newpage
\appendix
\onecolumn

\section{Proof of $k$-layer GNN output}
\subsection{$A^p$ denotation}\label{pf:Ap}

Proof of Lemma \ref{lm:Ap>0}.
\begin{proof}
We prove the statement by mathematical induction.

Let \( A \) be the adjacency matrix of a graph, where the elements \( A_{ij} \) are either \( 0 \) or positive integers. For any \( A_{ij} = 1 \), this indicates there is a directed edge from node \( i \) to node \( j \).

\textbf{Base Case:} For \( p = 1 \), \( A^p_{ij} = A_{ij} \). This holds true because \( A_{ij} > 0 \) if and only if there is a directed edge of length \( 1 \) from node \( i \) to node \( j \).

\textbf{Induction Hypothesis:} Assume that for some \( p \geq 1 \), a non-zero element \( A^p_{ij} > 0 \) indicates that there are exactly \( A^p_{ij} \) directed paths of length \( p \) from node \( i \) to node \( j \).

\textbf{Inductive Step:} For \( p + 1 \), consider the matrix multiplication \( A^{p+1} = A^p \cdot A \). By the rules of matrix multiplication, the element \( A^{p+1}_{ij} \) is given by:
\[
A^{p+1}_{ij} = \sum_{k} A^p_{ik} \cdot A_{kj}.
\]
Each term in the summation corresponds to the number of paths of length \( p \) from node \( i \) to node \( k \) (\( A^p_{ik} \)) multiplied by the number of paths of length \( 1 \) from node \( k \) to node \( j \) (\( A_{kj} \)).

Since \( A^p_{ik} \) and \( A_{kj} \) are non-negative integers, the summation \( \sum_{k} A^p_{ik} \cdot A_{kj} \) computes the total number of directed paths of length \( p+1 \) from node \( i \) to node \( j \). Therefore, \( A^{p+1}_{ij} > 0 \) if and only if there exists at least one directed path of length \( p+1 \) from node \( i \) to node \( j \).

Thus, by induction, the statement holds for all \( p \geq 1 \), and the entries of \( A^p \) represent the number of directed paths of length \( p \) between the corresponding nodes.
\end{proof}

\label{prf:k-lay}

\subsection{GCN with self-loop}
\label{proof:gcn-loop}
Proof of Lemma \ref{lm:gcn_loop}
Its layer-wise propagation is:
\begin{equation}
\label{rule:gcn}
    H^{(k+1)} = \sigma\big(W \odot (A+I) H^{(k)}W^{(k)} \big)
\end{equation}


\begin{proof}
\label{proof}

\textbf{Base Case:} 
In a 1-layer GCN,  the output is expressed as:
\[
 H^{(1)} = \sigma\big(W \odot (A+I) XW^{(1)} \big)
\]
which is consistent with the desired form.

\textbf{Inductive Step:} 
Assume that the statement holds for \(n=k\), i.e., 
\[
 H^{(k)}  = \sigma\big((W \odot (A+I))^k XW^{(k)} \big)
\]

For \(n=k+1\), according to layer-wise propagation rule in Eq. \ref{rule:gcn}:
\begin{equation}
\begin{split}
    H^{(k+1)} &= \sigma\big(W \odot (A+I) H^{(k)}\widehat W^{(k+1)} \big) \\
    &= \sigma\big(W \odot (A+I) \sigma((W \odot (A+I))^k XW^{(k)} )\widehat W^{(k+1)} \big) 
    \label{eq:k+1_gcnloop}
\end{split}
\end{equation}

Thanks to the Universal Approximation Theorem (UAT) \cite{HornikKurt1989Mfna, hornikApproximationCapabilitiesMuitilayer}, Eq.~\ref{eq:k+1_gcnloop} can be further simplified as:  
\begin{equation}
\begin{split}
    H^{(k+1)} &= \sigma\big(W \odot (A+I)(W \odot (A+I))^k XW^{(k)}\widehat W^{(k+1)} \big)\\
    &= \sigma\big((W \odot (A+I))^{(k+1)} XW^{(k+1)}\big)
\end{split}
\end{equation}
This simplification does not affect the network's ability to approximate the target function.

Thus, the statement holds for \(n=k+1\).

\textbf{Conclusion:}  
By the principle of mathematical induction, we conclude that for all \(n \geq 1\),
\[
  H^{(n)}  = \sigma\big((W \odot (A+I))^n XW^{(n)} \big)
\]
\end{proof}

\subsection{GCN without self-loop}
\label{prf:gcn_no_loop}
Proof of Lamma \ref{lm:gcn_no_loop}.

Its layer-wise propagation rule is:
\begin{equation}
\label{rule:gcn_no}
    H^{(k+1)} = \sigma\big(W \odot A H^{(k)}W^{(k)} \big)
\end{equation}


The proof follows similarly to Sec. \ref{proof:gcn-loop}, with $W \odot (A+I)$ replaced by $W \odot A$.

\subsection{GraphSAGE}
\label{prf:sage}
Proof of Lemma \ref{lm:sage}

Its layer-wise propagation rule is:
\begin{equation}
\label{rule:sage}
    H^{(k+1)} = \sigma\big( W \odot A H^{(k)}W_1^{(k)} + H^{(k)}W_0^{(k)} \big) 
\end{equation}


\begin{proof}
\textbf{Base Case:} 
In a 1-layer GCN,  the output is expressed as:
\[
 H^{(1)} = \sigma\big( W \odot A H^{(0)}W_1^{(1)} + H^{(0)}W_0^{(1)} \big) = \sigma\big( W \odot A XW_1^{(1)} + XW_0^{(1)} \big)
\]
which is consistent with the desired form.

\textbf{Inductive Step:} 
Assume that the statement holds for \(n=k\), i.e., 
\[
 H^{(k)}  = \sigma\big((W \odot A)^k XW_k^{(k)} + (W \odot A)^{(k-1)} XW^{(k)}_{k-1} +...+(W \odot A) XW_1^{(k)}+ XW^{(k)}_0\big)
\]

For \(n=k+1\), according to layer-wise propagation rule in Eq. \ref{rule:sage}:
\begin{equation}
\label{sage_k+1}
\begin{split}
    H^{(k+1)} = \sigma\Big( &W \odot A \cdot \sigma\big((W \odot A)^k XW_k^{(k)} + (W \odot A)^{(k-1)} XW^{(k)}_{k-1} +...+(W \odot A) XW_1^{(k)}+ XW^{(k)}_0\big)W_1^{(k)} \\
    &+ \sigma\big((W \odot A)^k XW_k^{(k)} + (W \odot A)^{(k-1)} XW^{(k)}_{k-1} +...+(W \odot A) XW_1^{(k)}+ XW^{(k)}_0\big)W_0^{(k)} \Big).
\end{split}
\end{equation}
Thanks to the Universal Approximation Theorem (UAT) \cite{HornikKurt1989Mfna, hornikApproximationCapabilitiesMuitilayer}, Eq.~\ref{sage_k+1} can be further simplified as: 
\begin{equation}
\begin{split}
    H^{(k+1)}  & = \sigma\Big(W \odot A \cdot \big((W \odot A)^k XW_k^{(k)} + (W \odot A)^{(k-1)} XW^{(k)}_{k-1} +...+(W \odot A) XW_1^{(k)}+ XW^{(k)}_0\big)W_1^{(k)} \\
    &+ \big((W \odot A)^k XW_k^{(k)} + (W \odot A)^{(k-1)} XW^{(k)}_{k-1} +...+(W \odot A) XW_1^{(k)}+ XW^{(k)}_0\big)W_0^{(k)} \Big) \\
    &= \sigma\Big(\big((W \odot A)^{k+1} XW_k^{(k)} + (W \odot A)^{k} XW^{(k+1)}_{k}  
    + ...+(W \odot A) XW_1^{(k+1)}+ XW^{(k)}_0\big)W_0^{(k+1)} \Big) \\
    &= \sigma\Big(\big((W \odot A)^{k+1} XW_{k+1}^{(k+1)} + (W \odot A)^{k} XW^{(k+1)}_{k}  
    + ...+(W \odot A) XW_1^{(k+1)}+ XW^{(k)}_0\big)W_0^{(k+1)} \Big)
\end{split}
\end{equation}
This simplification does not affect the network's ability to approximate the target function.

Thus, the statement holds for \(n=k+1\).

\textbf{Conclusion:}  
By the principle of mathematical induction, we conclude that for all \(n \geq 1\),
\[
  H^{(n)}  = \sigma\big((W \odot (A+I))^n XW^{(n)} \big)
\]
\end{proof}

\section{Proof of Loops' Influence}

\subsection{Proof of Self-loop Influence}
\label{pf:sloop}

Proof of Lemma \ref{lm:sloop}.
\begin{proof}
    Let node \( u \) be a \( k \)-hop neighbor of node \( v \). This implies that there exists a path from node \( u \) to node \( v \) consisting of \( k \) edges.

Each node in the graph has a self-loop, i.e., each node is connected to itself. This means that node \( v \) now has a direct edge to itself, in addition to the edges in the original graph.
    
    Since \( u \) was a \( k \)-hop neighbor of \( v \), the path from \( u \) to \( v \) that previously consisted of \( k \) hops can now include an additional hop from \( v \) to itself, forming a path of length \( k+1 \). Therefore, with the addition of self-loops, node \( u \), which was previously a \( k \)-hop neighbor of node \( v \), becomes a \( (k+1) \)-hop neighbor of node \( v \).
\end{proof}

\subsection{Proof of Two-node Loop Influence}
\label{pf:two_loop}

Proof of Lemma \ref{lm:two_loop}
\begin{proof}
Let node \( u \) be a \( k \)-hop neighbor of node \( v \). This implies that there exists a path from node \( u \) to node \( v \) consisting of \( k \) edges.
    
Since the graph is undirected, we can traverse any edge in both directions. Therefore, we can creates a valid \( (k+2) \)-hop path from \( v \) to \( u \) by:
- First following the original \( k \)-hop path
- Then moving backward along any edge
- Finally moving forward along the same edge

Therefore, the \( k \)-hop neighbors of node \( v \) become \( (k+2) \)-hop neighbors in the undirected graph.
\end{proof}

\subsection{Proof of Multi-node Loop Influence}
\label{pf:mul_loop}

Proof of Lemma \ref{lm:mul_loop}.
\begin{proof}
Consider a \( k \)-hop path \( P \) from node \( u \) to node \( v \). By our assumption, this path contains at least one node that is part of the \( m \)-length loop.

Let \( v_i \) be any such node on the loop that our path \( P \) passes through. We can construct a new \( (k+m) \)-hop path from \( v \) to \( u \) as follows:
\begin{itemize}
    \item Follow the original path from \( u \) to \( v_i \)
    \item At \( v_i \), take a full loop (which takes \( m \) steps and returns to \( v_i \))
    \item Continue on the original path from \( v_i \) to \( v \)
\end{itemize}

This new path reaches the same destination but uses exactly \( m \) more hops than the original path, making \( u \) a \( (k+m) \)-hop neighbor of \( v \).
\end{proof}

\subsection{Proof of Longest Path Influence for Directed Graph}
\label{pf:long_dire}
Proof of Lemma \ref{ls:directed_largest_hop}.

\begin{proof} 
1) In a directed graph without loops, every path is a simple path (no node can be visited twice).

2) Therefore:
   \begin{itemize}
   \item Any path longer than \( h \) would require visiting at least one node twice
   \item This is impossible as there are no loops in the graph
   \end{itemize}

3) Thus, there cannot exist any path of length \( m > h \), which means:
   \begin{itemize}
   \item All entries in \( A^m \) must be zero
   \item As \( A^m_{ij} \) counts the number of \( m \)-hop paths from \( i \) to \( j \)
   \end{itemize}
\end{proof}

\subsection{Proof of Longest Path Influence for Undirected Graph}
\label{pf:long_undire}

Proof of Lemma \ref{ls:undirected_largest_hop}.
\begin{proof}
We'll prove both directions of the equivalence:

1) First, any connection in \( A^h \) must also exist in \( A^m \) because in an undirected graph:
  \begin{itemize}
  \item If nodes are connected by an h-hop path
  \item They can always reach each other in more steps by going back and forth on edges
  \end{itemize}

2) Conversely, if nodes are connected in \( A^m \) (for $m > h$):
  \begin{itemize}
  \item They must be connected by some path in the graph
  \item Since h is the longest path length
  \item There must exist a path between them of length $ \leq$ h
  \item Therefore they are also connected in \( A^h \)
  \end{itemize}

Thus, the connections in \( A^m \) and \( A^h \) must be identical.
\end{proof}

\section{Datasets and Experimental Settings}

\subsection{Datasets}
\label{ap:data}
We use six widely-adopted real-world datasets, all of which are directed graphs, comprising three citation networks (CiteSeer, CoraML, PubMed), two webpage traffic classification (Chameleon, Squirrel), one social network (Telegram).
Dataset statistics are reported in Table \ref{tab:statis}.
\begin{table*}[ht]
    \centering
    \captionsetup{font=small}
    \caption{Dataset statistics. Imbalance Ratio is the ratio of the largest class to the smallest class in the training sets. \%No-In and \%No-Out represent the percentage of nodes with no direct in-neighbors and no direct out-neighbors, respectively. \%In-Homo denotes the percentage of nodes whose in-neighbors predominantly share the same label as the node, while \%Out-Homo indicates the percentage of nodes whose out-neighbors predominantly share the same label.}
   \fontsize{9}{11}\selectfont
{
    \begin{tabular}{c|ccccc|ccccc}
    \toprule
Dataset & \#Nodes & \#Edges &\#Feat. &  \#C & Imbal-Ratio & \%No-In & \%In-Homo & \%No-Out & \%Out-Homo  & Label rate\\
CiteSeer &3312 &4715 &3703 & 6 & 1.0 & 30.2 & 52.7 & 41.1 & 44.1 &  3.6\% \\
Cora-ML  & 2995&8416 & 2879&7 &1.0 &  41.7 & 50.2 & 11.7 & 74.7 &  4.8\% \\

PubMed &19717 &44327 &500 &3 &1.0 &  33.4 & 54.2 & 34.2 & 54.0 &  0.3\%\\
\midrule
Chameleon & 2277 & 36101 & 2325& 5 & 1.3 & 62.1 & 10.4 & 0.0 & 25.3 & 48\%  \\
Squirrel &5201 & 217073&2089 & 5 & 1.1&  57.6 & 8.5 & 0.0 & 24.2 & 48\% \\
Telegram & 245& 8912 & 1&4 & 3.0 &  16.7 & 32.2 & 25.3 & 32.7 &  60\% \\
\bottomrule     

    \end{tabular}}

    \label{tab:statis}
\end{table*}

We categorize these datasets into two types based on how graph neural networks (GNNs) utilize their information:

\paragraph{Structure-feature Hybrid Type}
The following citation network datasets combine both structural information (edges) and node features. For all these datasets, we use 10 random splits with 20 nodes per class for training, 30 nodes per class for validation, and the remaining nodes for testing.
\begin{itemize}
    \item \textbf{CiteSeer} and \textbf{CoraML} are
    academic citation networks where nodes represent research papers and edges represent citations between papers. Node features are derived from document content, and classes correspond to different research areas.
   \item \textbf{PubMed} is a citation network of medical research papers, obtained from the Deep Graph Library (DGL). We generate 10 random splits following the same protocol as CiteSeer and CoraML.

\end{itemize}

\paragraph{Structure-only Type}
These datasets demonstrate the effectiveness of GNNs in scenarios where structural connection play a dominant role:
\begin{itemize}
        \item \textbf{Chameleon and Squirrel} are Wikipedia page networks where nodes represent articles and edges represent hyperlinks between pages \cite{rozemberczki2021multi}. Node features are derived from the frequency of specific nouns in the articles. The classification task involves predicting the average monthly traffic volume for each page, categorized into five levels. We adopt the data splits established in GEOM-GCN \cite{peiGeomGCNGeometricGraph2020}.

\item \textbf{Telegram} represents a complex interaction network within the Telegram platform, encompassing relationships between channels, chats, posts, and URLs. We follow the experimental setup and data splits defined in MagNet \cite{zhangMagNetNeuralNetwork}.
    \end{itemize}

\subsection{Experimental Settings}
\subsubsection{Implementation Details}
All experiments are carried on NVIDIA A40 GPUs with 48GB VRAM.

For experiments involving Chameleon and Squirrel datasets, we followed the methodology outlined by Rossi et al. \cite{rossiEdgeDirectionalityImproves2023}, using a learning rate of 0.005, a patience of 400 epochs, and a total training duration of 100,000 epochs, with early stopping applied after 810 epochs of no improvement in validation accuracy.

For other datasets,we employed the Adam optimizer and trained the model for 1,500 epochs, utilizing early stopping based on validation accuracy with a patience of 410 epochs. Additionally, a learning rate scheduler was applied, with a patience of 80 epochs. 
We report the mean and standard deviation of test accuracy over 10 runs.

\subsubsection{Hyperparameter Tuning}

For each model, we perform a grid search to optimize the following hyperparameters:

\begin{itemize}
    \item \textbf{Learning Rate:} 0.1, 0.01, 0.005
    \item \textbf{L2 Regularization}: 0, 0.0005
    \item \textbf{Dropout Rate:} 0.0, 0.5
\end{itemize}

\section{Additional Experiments}
\label{ap_add_experi}

\subsection{Experiments on Reverse Direction and Bidirectional Propagation}
\begin{figure}
\centering
\captionsetup{font=small}

\begin{subfigure}{0.45\textwidth}
\includegraphics[width=\linewidth]{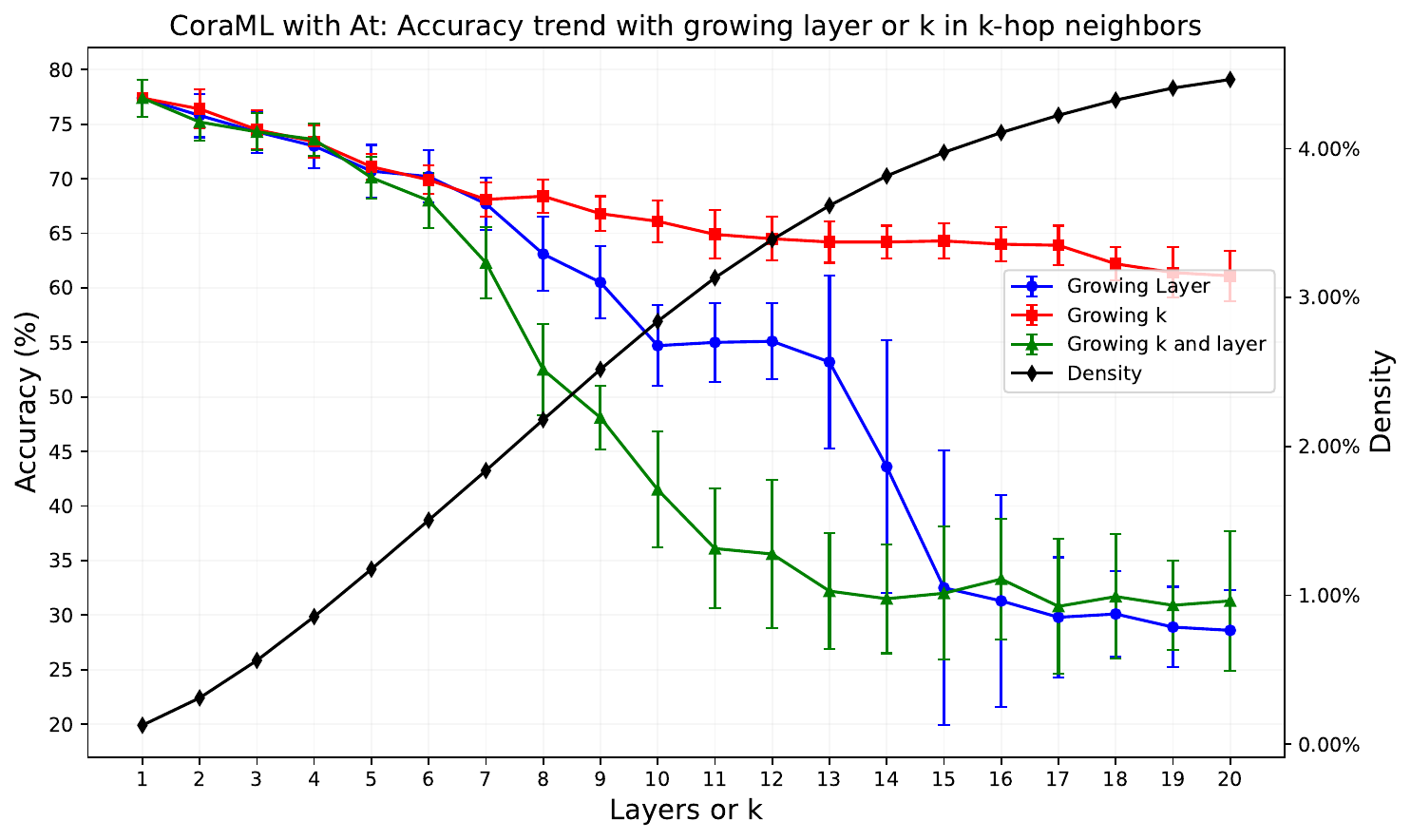}
\caption{$A^T$ as adjacency matrix}
\label{fig_cora_At}
\end{subfigure}
\hfill
\begin{subfigure}{0.45\textwidth}
\includegraphics[width=\linewidth]{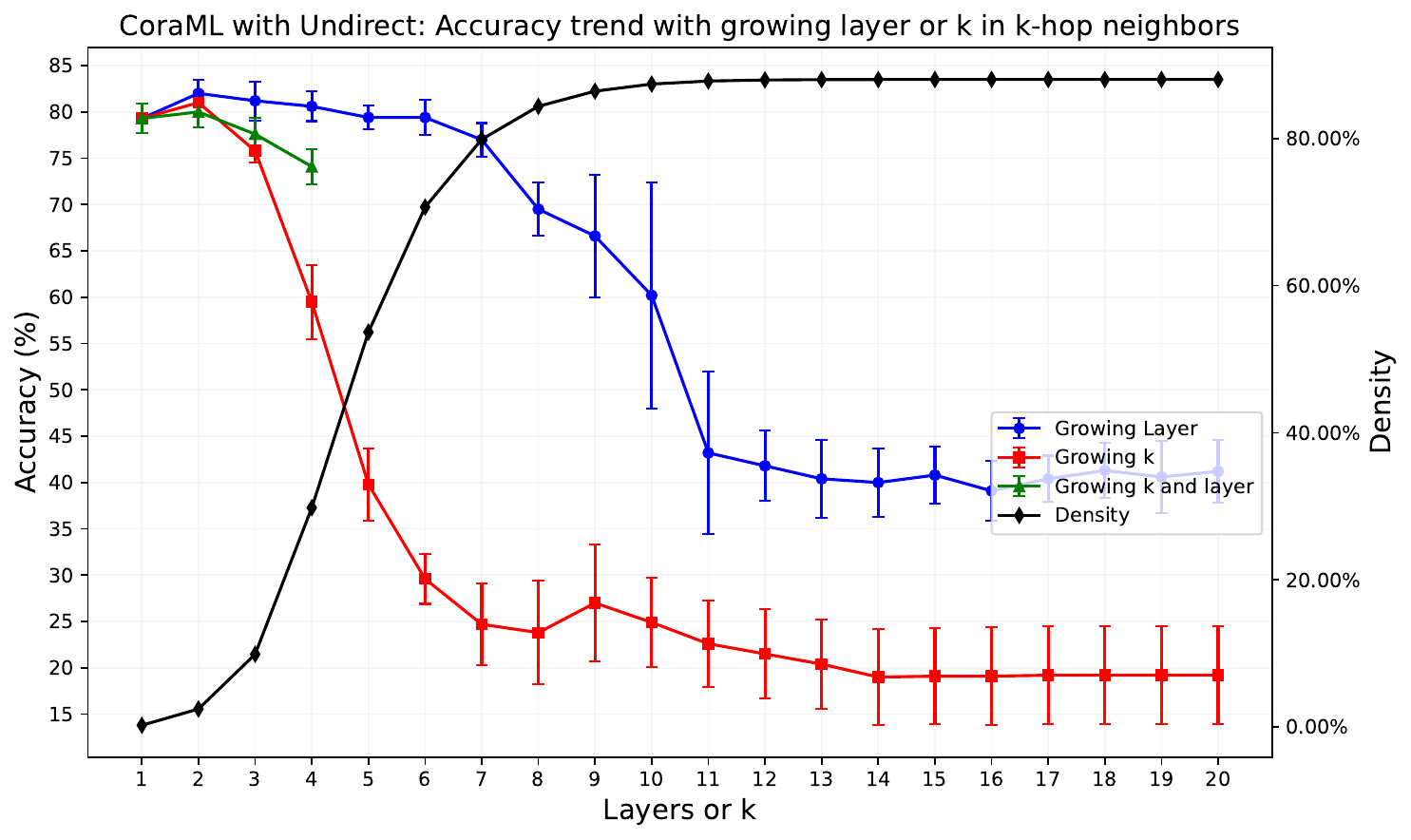}
\caption{Bidirectional Propagation}
\label{fig_cora_undirect}
\end{subfigure}
\vskip -0.15in
\caption{Comparison of different GCN architectures on CoraML dataset under different adjacency matrix formulations. (Top) Using transposed adjacency matrix $A^T$, which propagates information from cited papers to citing papers. (Bottom) Using undirected graph adjacency matrix $A + A^T$, which enables bidirectional information flow. In each subplot: k-layer GCN (blue), 1-layer GCN with k-hop neighbors (red), and k-hop neighbors with (k-1) linear layers (green). The black line indicates the density of the k-hop adjacency matrix. Missing experimental results is due to out of memory on NVIDIA A40 GPUs with 48GB VRAM.}
\end{figure}
\begin{figure}
\centering
\captionsetup{font=small}
\begin{subfigure}{0.45\textwidth}
\includegraphics[width=\linewidth]{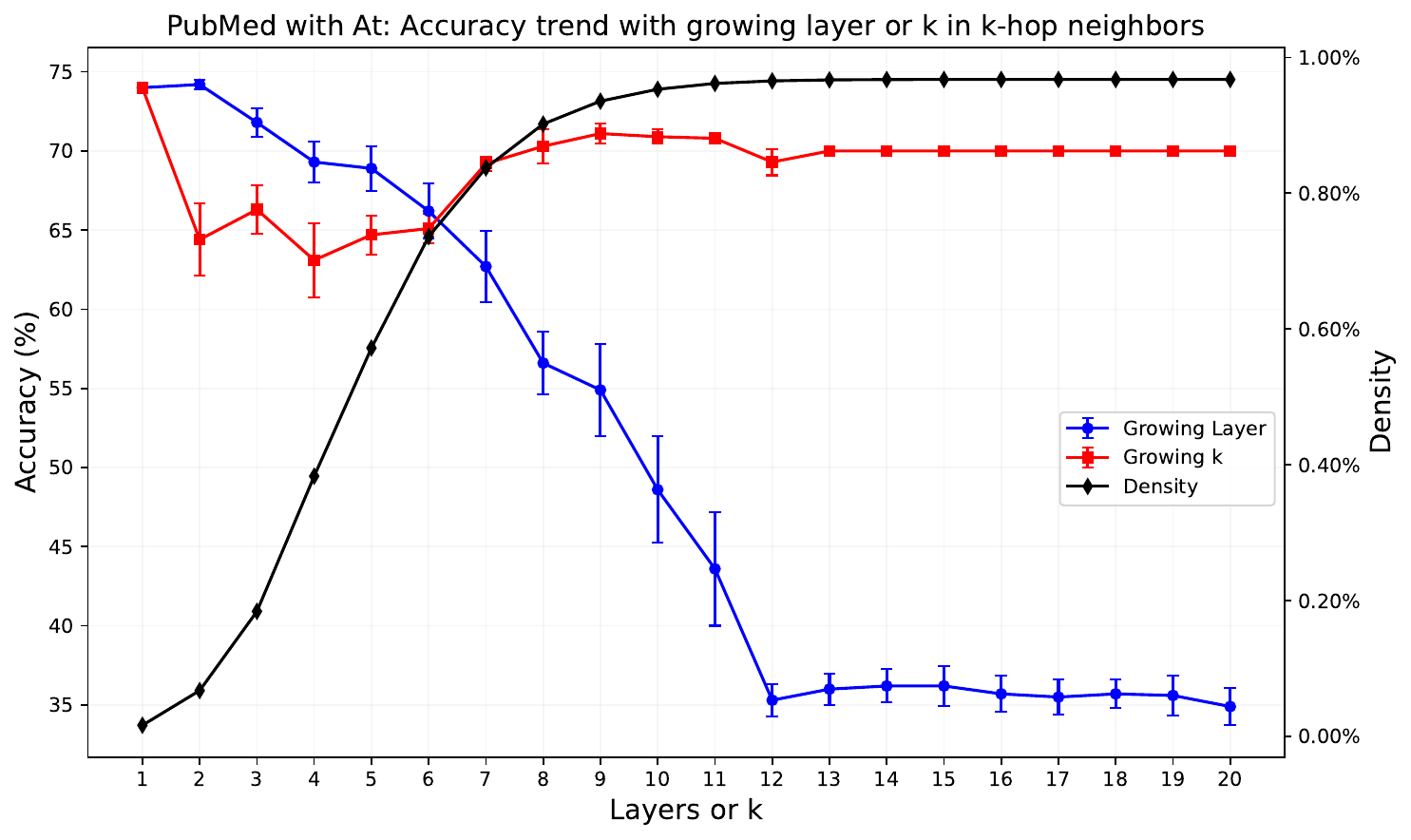}
\caption{$A^T$ as adjacency matrix}
\label{fig_pub_At}
\end{subfigure}
\hfill
\begin{subfigure}{0.45\textwidth}
\includegraphics[width=\linewidth]{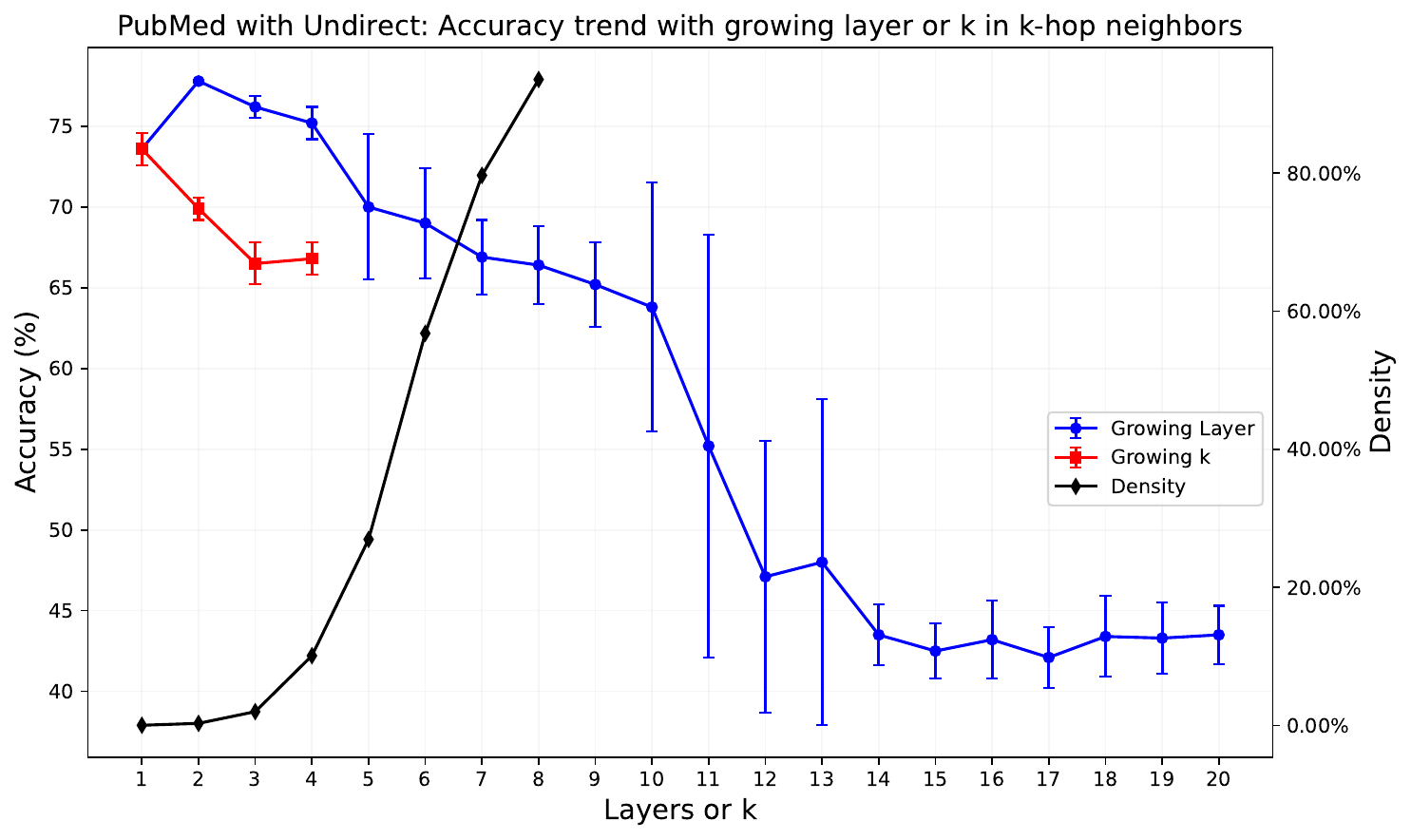}
\caption{Bidirectional Propagation}
\label{fig_pub_undirect}
\end{subfigure}

\vskip -0.15in
\caption{Comparison of different GCN architectures on PubMed dataset under different adjacency matrix formulations. (Top) Using transposed adjacency matrix $A^T$, which propagates information from cited papers to citing papers. (Bottom) Using undirected graph adjacency matrix $A + A^T$, which enables bidirectional information flow. In each subplot: k-layer GCN (blue), 1-layer GCN with k-hop neighbors (red), and k-hop neighbors with (k-1) linear layers (green). The black line indicates the density of the k-hop adjacency matrix. Missing experimental results is due to out of memory on NVIDIA A40 GPUs with 48GB VRAM.}
\end{figure}

\section{Literature Review}






\subsection{Graph Neural Networks}  
Graph Neural Networks (GNNs) extend Multi-Layer Perceptrons (MLPs) by incorporating a message-passing sublayer before the linear transformation and activation. Different GNN architectures primarily vary in their message-passing mechanisms.  

The predominant paradigm in GNNs is message passing. \citet{gilmerNeuralMessagePassing2017} formalized a general framework for supervised learning on graphs, known as Message Passing Neural Networks (MPNNs). Existing and newly developed GNNs can be viewed as variations of models within the MPNN family.  

GNNs leverage graph structure and node features $X_v$ to learn a representative vector for each node, $h_v$. Modern GNNs follow a neighborhood aggregation strategy, where a node's representation is iteratively updated by aggregating information from its neighbors. After $k$ iterations of aggregation, the node representation is expected to encode structural information within its $k$-hop neighborhood \cite{xuHowPowerfulAre2019}.  

A typical layerwise operation in GNNs consists of two steps: aggregation and combination. In the aggregation step, a node collects information from its 1-hop neighbors. Some specially designed models, such as those handling higher-order neighborhoods \cite{jiangScaleInvarianceGraph2024, tongDirectedGraphConvolutional2020}, extend this process to 2-hop or $k$-hop neighborhoods in each step.

GNN variants mostly differ in their choices of aggregation and combination. 
A number of architectures for GNN variants have been proposed.
Graph Convolutional Network (GCN) \cite{kipfSemiSupervisedClassificationGraph2017} aggregates and normalizes neighbor information. the layer-wise output has been formulated as 
\begin{equation}
\label{eq:gcn-ori}
    H^{(k+1)} = \sigma\big( \widetilde{A}_{\text{sym}} H^{(k)}W^{(k)} \big) 
\end{equation}
where $\widetilde{A}_{\text{sym}} = \widetilde{D}^{-\frac{1}{2}} {(A+I)} \widetilde{D}^{-\frac{1}{2}}$ is a symmetric normalized adjacency matrix with added self-loops, $\widetilde{D}$ is the degree matrix of $A+I$.  $H^{(k)}$ represents  features from the previous layer, $W^{(k)}$ denotes the learnable weight matrix, and $\sigma$ is a non-linear activation function. $H^0$ is the original node feature $X$.

As normalization is a process of adjusting weight for each element, Eq. \ref{eq:gcn-ori} can be further expressed as:
\begin{equation}
\label{eq:gcn_norm}
    H^{(k+1)} = \sigma\big(W \odot (A+I) H^{(k)}W^{(k)} \big)
\end{equation}
where $W$ is normalization weight matrix, and $\odot$ denotes element-wise multiplication. 

For heterophilic graphs, where nodes have dissimilar features and labels to their neighbors, avoiding self-loops can improve performance by reducing the aggregation of dissimilar features \cite{zhu2020beyond, rossiEdgeDirectionalityImproves2024}.
\begin{equation}
\label{eq:no-loop}
    H^{(k+1)} = \sigma\big(W \odot A H^{(k)}W^{(k)}\big)
\end{equation}

GraphSAGE \cite{hamiltonInductiveRepresentationLearning2018} concatenates self-node features with its neighbors' features.
\begin{equation}
    H^{(k+1)} = \sigma\big( W \odot A H^{(k)}W_1^{(k)} + H^{(k)}W_0^{(k)} \big) 
\end{equation}

In these networks, at each layer, each node updates its representation as follows \cite{bechler-speicherGraphNeuralNetworks2024}:
\begin{equation}
h_i^{(k+1)} = \sigma \left( W_1^{(k)} h_i^{(k)} + W_2^{(k)} \sum_{j \in \mathcal{N}(i)} h_j^{(k)} + b^{(k)} \right)
\label{eq:sum}
\end{equation}

\subsection{Message Passing Neural Networks}

\cite{gilmerNeuralMessagePassing2017} introduced the Message Passing Neural Network (MPNN) framework, which serves as a general paradigm for supervised learning on graphs. While newer Graph Neural Networks (GNNs) achieve better performance, they remain variations within the MPNN family.

GNNs leverage both graph structure and node features $X_v$ to learn node representations $h_v$. Modern GNNs employ a neighborhood aggregation strategy, where a node iteratively updates its representation by aggregating information from its neighbors. After $k$ iterations, the representation is believed to capture structural information within the node’s $k$-hop neighborhood \cite{xuHowPowerfulAre2019}. A typical GNN layer consists of an aggregation step, where 1-hop neighbors are gathered, followed by a combination step. Some specialized models incorporate higher-order neighborhoods, aggregating 2-hop or even $k$-hop neighbors in a single step \cite{jiangScaleInvarianceGraph2024, tongDirectedGraphConvolutional2020}. GNN variants primarily differ in their choices of aggregation and combination functions.

While GNNs integrate both structural and feature information for predictions, how these components interact remains unclear. It is commonly assumed that a $k$-layer GNN effectively synthesizes structural and feature information to obtain comprehensive node representations. However, our research reveals that iterative neighborhood aggregation is equivalent to replacing $k$-hop neighbors with $(k+1)$-hop neighbors. The introduction of self-loops fundamentally alters this process: iterative aggregation then becomes equivalent to progressively incorporating $(k+1)$-hop neighbors atop the existing 1-hop, 2-hop, ..., $k$-hop neighbors. 

Our findings highlight the need for a deeper theoretical understanding of how neighborhood aggregation impacts representational expressiveness in GNNs.

\subsection{Over-smoothing}

\paragraph{Definition}  
Many studies have empirically identified that GNNs tend to smooth node representations as layers deepen, ultimately causing them to become nearly indistinguishable—an issue known as over-smoothing \cite{ruschSurveyOversmoothingGraph2023}. This results in a decline in model performance due to the loss of distinguishable node features.  

Over-smoothing occurs when node representations become increasingly similar, whereas over-squashing \cite{stanovicGraphNeuralNetworks2025} arises when long-range information is compressed, leading to significant information loss. Over-squashing specifically refers to the inability of a GNN to effectively transfer information between distant nodes.  

\paragraph{Causes}  
The over-smoothing problem \cite{mauryaSimplifyingApproachNode2022} in GNNs is often attributed to excessive feature averaging over multiple hops, which reduces the informativeness of node features. \citet{mauryaSimplifyingApproachNode2022} suggest that addressing over-smoothing can be framed as a feature selection problem.  

Moreover, excessive aggregation can be redundant or even detrimental, ultimately causing all node representations to converge to a common fixed point \cite{xieWhenGNNsWork2020}.  

\paragraph{Mitigation Strategies}  
Various strategies have been proposed to mitigate over-smoothing and over-squashing. \citet{stanovicGraphNeuralNetworks2025} explore the use of pooling operators as a potential solution.  

GRAND \cite{chamberlainGRANDGraphNeural2021a} interprets GNNs as a discretization of a heat diffusion equation, highlighting that learning distinguishable node representations in deep GNNs is inherently difficult, as heat diffusion naturally leads to equilibrium, making representations indistinguishable.  

For heterophilic graphs, \citet{parkMitigatingOversmoothingReverse2024} propose a reverse aggregation process to counteract the diffusive nature of GNNs. This approach enables the effective stacking of hundreds or even thousands of layers, thereby mitigating over-smoothing and facilitating the capture of long-range dependencies—an essential capability for heterophilic datasets.

\end{document}